\def\paperID{0000}
\def\confName{ICCV}
\def\confYear{2025}

\def\paperTitle{SemanticSplat: Feed-Forward 3D Scene Understanding with Language-Aware Gaussian Fields}

\def\authorBlock{
    Qijing Li\thanks{Equal contribution} \qquad
    Jingxiang Sun\footnotemark[1] \qquad
    Liang An \qquad
    Zhaoqi Su \qquad
    Hongwen Zhang \qquad
    Yebin Liu\thanks{Corresponding author} \\
    \\
    Tsinghua University \qquad
    Beijing Normal University
}

\newif\ifreview 
\newif\ifarxiv \newcommand{\arxiv}{\arxivtrue}
\newif\ifcamera 
\newif\ifrebuttal 

\arxiv 

\pdfoutput=1
\documentclass[10pt,twocolumn,letterpaper]{article}
\ifreview \usepackage[review]{cvpr} \fi
\ifarxiv \usepackage[pagenumbers]{cvpr} \fi
\ifrebuttal \usepackage[rebuttal]{cvpr} \fi
\ifcamera \usepackage{cvpr} \fi

\usepackage{graphicx}	
\usepackage{amsmath}	
\usepackage{amssymb}	
\usepackage{booktabs}
\usepackage{times}
\usepackage{microtype}
\usepackage{epsfig}
\usepackage{caption}
\usepackage{float}
\usepackage{placeins}
\usepackage{color, colortbl}
\usepackage{stfloats}
\usepackage{enumitem}
\usepackage{tabularx}
\usepackage{xstring}
\usepackage{multirow}
\usepackage{xspace}
\usepackage{url}
\usepackage{subcaption}
\usepackage{xcolor}
\usepackage[hang,flushmargin]{footmisc}
\usepackage{hyperref}
\usepackage{xspace}
\usepackage{footmisc}
\usepackage{tikz}
\usepackage{adjustbox}
\usepackage{subcaption}
\usepackage[most]{tcolorbox} 
\usepackage{stfloats}

\ifcamera \usepackage[accsupp]{axessibility} \fi

% ...

\ifarxiv  \fi

\newcommand{\R}[1]{{%
    \textbf{%
        \ifstrequal{#1}{1}{\textcolor{red}{R#1}}{%
        \ifstrequal{#1}{2}{\textcolor{blue}{R#1}}{%
        \ifstrequal{#1}{3}{\textcolor{magenta}{R#1}}{%
        \ifstrequal{#1}{4}{\textcolor{teal}{R#1}}{%
                           \textcolor{cyan}{R#1}%
        }}}}%
    }%
}}

\usepackage{xr-hyper}

\makeatletter
\newcommand*{\addFileDependency}[1]{
  \typeout{(#1)}
  \@addtofilelist{#1}
  \IfFileExists{#1}{}{\typeout{No file #1.}}
}

\makeatother
\newcommand*{\myexternaldocument}[1]{
    \externaldocument{#1}
    \addFileDependency{#1.tex}
    \addFileDependency{#1.aux}
}

\definecolor{cvprblue}{rgb}{0.21,0.49,0.74}
\usepackage[capitalize]{cleveref}
\crefname{section}{Sec.}{Secs.}
\crefname{table}{Table}{Tables}
\crefname{figure}{Fig.}{Figs.}

\ifarxiv \crefname{appendix}{App.}{Apps.}
\else \crefname{appendix}{Suppl.}{Suppls.} \fi

\unless\ifarxiv \myexternaldocument{_supplementary} \fi

\begin{document}
\title{\paperTitle}
\author{\authorBlock}

\twocolumn[{%
\maketitle
\begin{figure}[H]
\hsize=\textwidth 
\centering
\includegraphics[width=\textwidth]{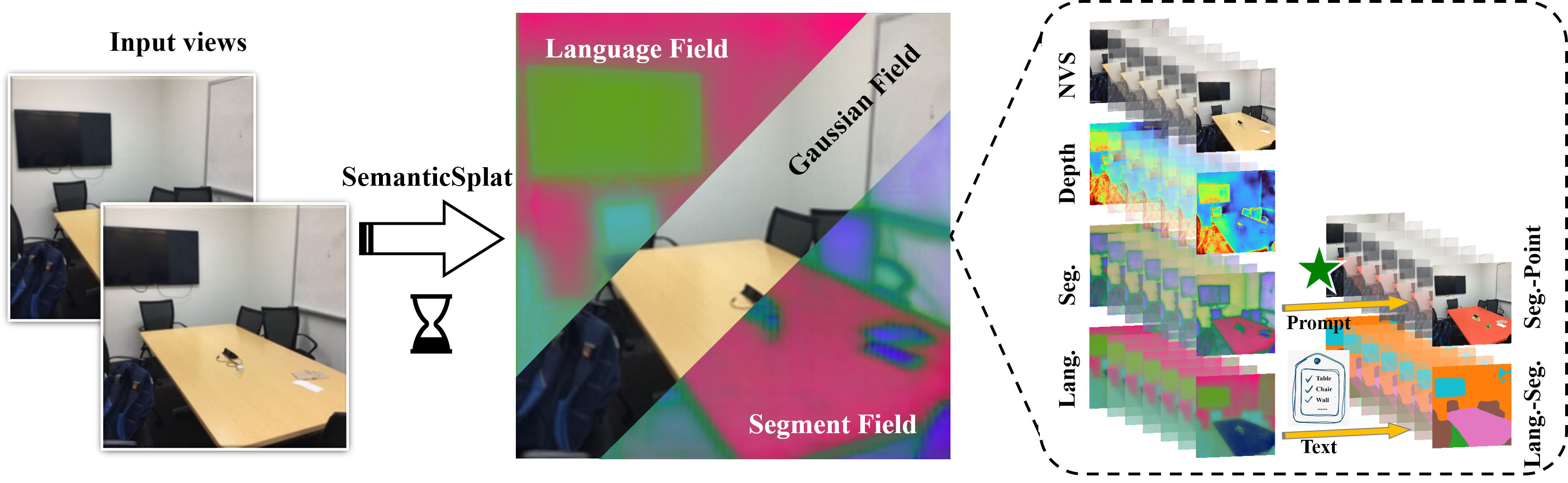}
\caption{Our approach utilizes sparse view images as input to reconstruct a holistic semantic Gaussian field, which includes both the Gaussian field with language features and the segmentation features. This reconstruction captures geometry, appearance, and multi-modal semantics, enabling us to perform multiple tasks such as novel view synthesis, depth prediction, open-vocabulary segmentation, and promptable segmentation.}
\end{figure}
}]

\newcommand{\ourpaper}{SemanticSplat\xspace}

\begin{abstract}
Holistic 3D scene understanding, which jointly models geometry, appearance, and semantics, is crucial for applications like augmented reality and robotic interaction. 
Existing feed-forward 3D scene understanding methods (e.g., LSM) are limited to extracting language-based semantics from scenes, failing to achieve holistic scene comprehension. Additionally, they suffer from low-quality geometry reconstruction and noisy artifacts. In contrast, per-scene optimization methods rely on dense input views, which reduces practicality and increases complexity during deployment. 
In this paper, we propose \ourpaper, a feed-forward semantic-aware 3D reconstruction method, which unifies 3D Gaussians with latent semantic attributes for joint geometry-appearance-semantics modeling. To predict the semantic anisotropic Gaussians, \ourpaper fuses diverse feature fields (e.g., LSeg, SAM) with a cost volume representation that stores cross-view feature similarities, enhancing coherent and accurate scene comprehension. Leveraging a two-stage distillation framework, \ourpaper reconstructs a holistic multi-modal semantic feature field from sparse-view images.
Experiments demonstrate the effectiveness of our method for 3D scene understanding tasks like promptable and open-vocabulary segmentation. Video results are available at https://semanticsplat.github.io.
\end{abstract}
\section{Introduction}
The ability to achieve holistic 3D understanding from 2D imagery is central to many applications in robotics, augmented reality (AR), and interactive 3D content creation. Such tasks demand representations that seamlessly combine precise geometry, realistic appearance, and flexible semantics. Traditional pipelines typically decompose this goal into multiple distinct stages: Structure‑from‑Motion (SfM) for sparse camera pose estimation, Multi‑View Stereo (MVS) for dense geometry recovery, and specialized modules for semantic labeling. Although effective in structured scenarios, this staged approach is prone to error propagation—small inaccuracies in early stages (e.g., pose estimation) often amplify through subsequent steps, resulting in degraded semantic and geometric reconstructions. Moreover, reliance on dense, accurately calibrated views severely restricts applicability in less controlled, real‑world environments. Additionally, the lack of extensive labeled 3D datasets limits these methods’ ability to generalize beyond fixed semantic categories, hampering open‑vocabulary scene understanding.

Recently, 3D scene understanding methods leveraging powerful pre‑trained 2D foundational models—such as the Segment Anything Model (SAM)~\cite{SAM} and CLIP~\cite{CLIP}—has emerged as a paradigm to enrich 3D representations with semantic knowledge distilled from readily available 2D data. However, directly transferring 2D semantic knowledge to 3D is non-trivial: 2D predictions often suffer from view-dependent inconsistencies, leading to noisy and unreliable semantic fields when aggregated across views. 
Besides, while NeRF~\cite{NeRF} and explicit 3D Gaussian splatting~\cite{3DGS} methods enhanced with 2D features have shown promise for open‑vocabulary flexibility, existing approaches predominantly depend on per-scene optimization, making them impractical for dynamic or large‑scale applications.

In this paper, we propose \ourpaper, a feed-forward framework for joint 3D reconstruction and semantic field prediction from sparse input images. Our approach extends 3D Gaussian Splatting by augmenting each Gaussian with latent semantic attributes, enabling simultaneous rendering of RGB and semantic feature maps. This unified representation jointly encodes geometry and semantics within a single framework, where the learned semantic attributes maintain multi-view consistency while preserving the efficiency of 3D Gaussian representations. By distilling knowledge from pre-trained visual foundation models (VFMs) like SAM and CLIP-LSeg, we achieve robust and accurate promptable and open-vocabulary segmentation.
Our key contributions include:
\begin{enumerate}
    \item \textbf{Feed-forward Holistic 3D Scene Understanding} – We propose a feed-forward semantic-aware method to predict semantic anisotropic Gaussians augmented with latent semantic features, enabling joint optimization of geometry, appearance, and multi-modal semantics. This facilitates a comprehensive understanding of 3D scenes.
    \item \textbf{Multi-Conditioned Feature Fusion} – We propose a novel pipeline that aggregates monocular semantic features (from SAM and CLIP-LSeg) with multi-view cost volumes, improving cross-view consistency and semantic awareness in complex scenarios.
    \item \textbf{Two-Stage Feature Distillation} – We separately lift SAM and CLIP-LSeg features into 3D through a two-stage process, reconstructing both segmentation and language feature fields. This supports multi-modal segmentation, including promptable and open-vocabulary segmentation.
\end{enumerate}
\section{Related Work}

\subsection{3D Scene Understanding}
Early language-aware scene representations embed CLIP features in NeRF volumes to support open-vocabulary queries, as demonstrated by LERF~\cite{lerf}. Subsequent efforts migrate to 3D Gaussian Splatting (3DGS) for real-time rendering: GARField~\cite{garfield} distills SAM masks into Gaussians, LangSplat~\cite{Langsplat} auto-encodes a scene-wise language field, and Gaussian Grouping~\cite{gaussian-grouping} attaches identity codes for instance-level clustering. Recent extensions such as SAGA~\cite{cen2025segment} and 4D LangSplat~\cite{li20254d} provide promptable segmentation and temporally coherent language fields, respectively, yet their reliance on point-cloud surfaces limits mesh fidelity. Emerging approaches like OV-NeRF~\cite{liao2024ov} introduce cross-view self-enhancement strategies to mitigate CLIP's view inconsistency through semantic field distillation, while DiCo-NeRF~\cite{choi2024dico} leverages CLIP similarity maps for dynamic object handling in driving scenes. Concurrent work MaskField~\cite{gao2024fast} demonstrates how decomposing SAM mask features from CLIP semantics enables efficient 3D segmentation in Gaussian Splatting representation. Although fast in per-scene training, it still needs per-scene optimization.

\subsection{Feed-Forward Gaussian Splatting}
Feed-forward reconstructors amortize 3DGS inference. For example, PixelSplat~\cite{pixelsplat} learns Gaussians from two views, while Splatter Image~\cite{szymanowicz2024splatter} accelerates single-view object recovery through per-pixel Gaussian prediction. The recent Hierarchical Splatter Image extension~\cite{shen2024pixel} introduces parent-child Gaussian structures to recover occluded geometry through view-conditioned MLPs. To leverage multi-view cues, MVSplat~\cite{mvsplat} builds cost volumes; we push this idea further by injecting monocular depth priors for texture-less scenes. Large-scale models trade hand-crafted geometry for data-driven priors: LGM~\cite{tang2024lgm}, GRM~\cite{xu2024grm}, GS-LRM~\cite{zhang2024gs}, and LaRA~\cite{chen2024lara} reconstruct scenes in milliseconds but demand >60 GPU-days for pre-training. Gamba~\cite{shen2024gamba} achieves 1000× speedup over optimization methods through Mamba-based sequential prediction of 3D Gaussians, though constrained to object-level reconstruction. Our approach reaches comparable quality in two GPU-days and, unlike LRMs, can be pre-trained with inexpensive posed images \emph{without} depth supervision.

\subsection{Lifting 2D Foundation Models to 3D}
Neural fields can aggregate multi-view image features into a canonical 3D space. Semantic NeRF~\cite{semantic-nerf} and Panoptic Lifting~\cite{panoptic-lifting} fuse segmentation logits, showing that consistent 3D fusion cleans noisy 2D labels. Beyond labels, Distilled Feature Fields~\cite{decomposing-nerf}, LERF~\cite{lerf2023}, NeRF-SOS~\cite{nerf-sos}, and FeatureNeRF~\cite{featurenerf} render pixel-aligned DINO or CLIP embeddings for tasks such as key-point transfer. Recent 3DGS adaptations~\cite{lee2025rethinking, liu2024splatraj, Langsplat, legaussian, li2024langsurf, zhou2024feature, fmgs, zhou2025feature4x, LSM} adopt similar strategies to distill information from well-trained 2D models to 3D Gaussians. Feature 3DGS~\cite{zhou2024feature} generalizes distillation to explicit Gaussians; concurrent works like FMGS~\cite{fmgs} and SPLAT-Raj~\cite{liu2024splatraj} confirm that SAM or LSeg signals can be attached to Gaussians for open-vocabulary editing.

\section{Method}
\begin{figure*}
	\centering
	\includegraphics[width=0.99\linewidth]{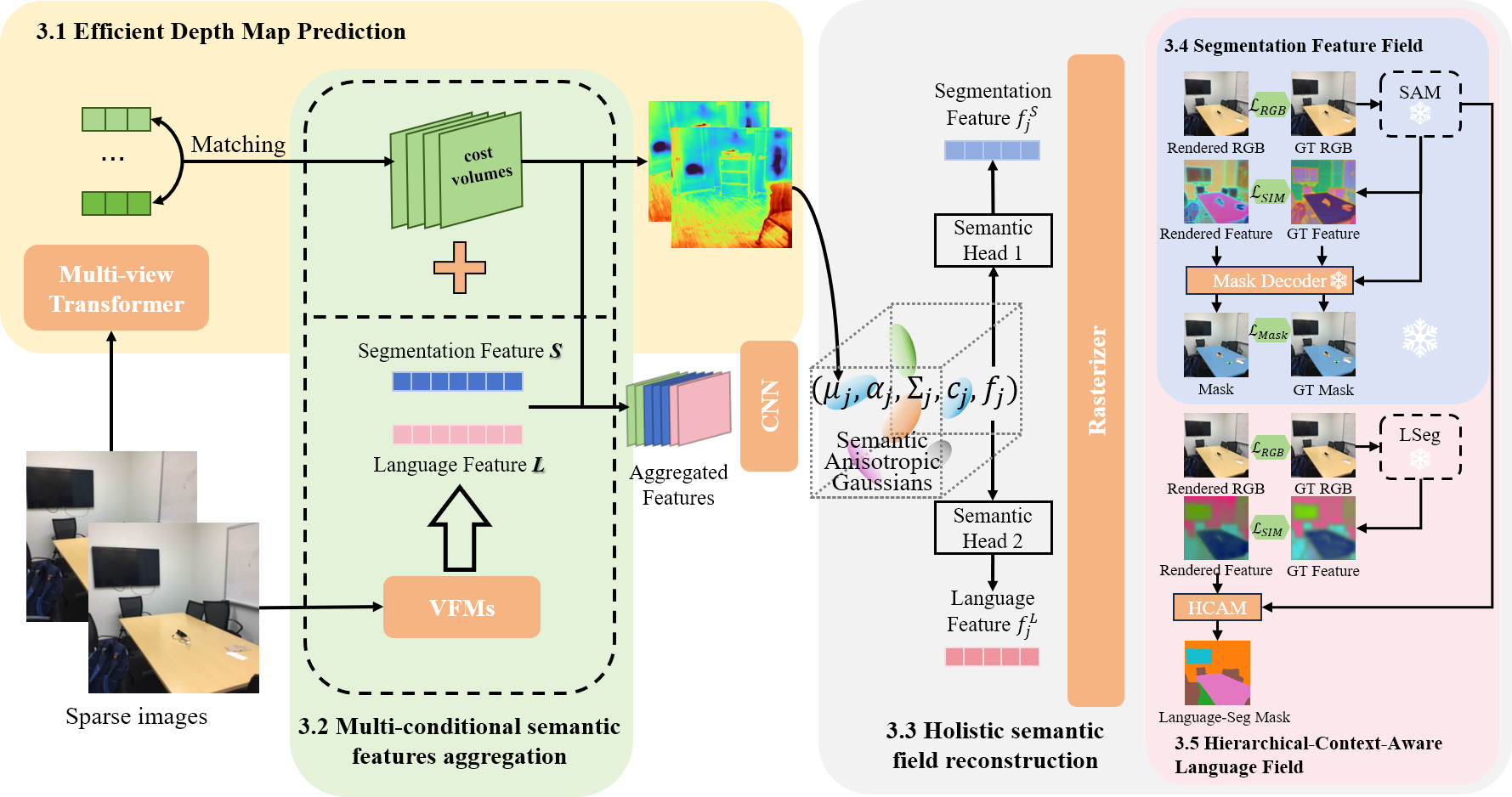}
	\caption{We employ multiview transformers with cross-attention to extract features from multi-view images and use cost volumes for feature matching (see Sec.~\ref{3.1}). Utilizing the multi-conditioned semantic features from Visual Feature Modules (VFMs) aggregated with the cost volumes (see Sec.~\ref{3.2}), we predict Semantic Anisotropic Gaussians(see Sec.~\ref{3.3}). Through a two-stage feature distillation process involving both segmentation(see Sec.~\ref{sec:sam_distill}) and language features(see Sec.~\ref{sec:lseg_distill}), we reconstruct the holistic semantic feature field by jointly enforcing photometric fidelity and semantic consistency. }
	\label{fig:pipeline}
\end{figure*}

\paragraph{Overview.} The goal of our \ourpaper is to holistically reconstruct the 3D scene with multi-modal semantics. As shown in Figure~\ref{fig:pipeline}, given N sparse input images${\left\{I_i\in{\mathbb{R}^{H\times{W}\times{3}}}\right\}_{i=1}^N}$, with associated camera projection matrices ${\left\{P_i=K_i\left[R_i\mid{T_i}\right]\right\}_{i=1}^N}$ 
we propose to predict per-pixel semantic anisotropic Gaussians \\
${\left\{({\mu_j, \alpha_j,\sum_j,c_j,f_j})\right\}_{j=1}^{H\times{W}\times{N}}}$ for each image, representing the holistic semantic features of the scene, including segmentation features and language-aligned features. This enables feed-forward novel view synthesis and multi-modal segmentation of the scene, including promptable segmentation and open-vocabulary segmentation.

The per-pixel Gaussians are predicted through ViT-based feature matching using cost volumes, with per-view depth maps regressed by a 2D U-Net (Sec.~\ref{3.1}). Inspired by DepthSplat, we propose a new branch that conditions on multi-source monocular semantic features (Sec.~\ref{3.2}) to enhance comprehension quality. In parallel with depth prediction, we introduce an auxiliary head to predict per-pixel semantic feature embeddings in 3D Gaussian space (Sec.~\ref{3.3}). Leveraging a two-stage feature distillation process, we reconstruct a holistic semantic field lifted from 2D pretrained models (Sec.~\ref{sec:sam_distill} and Sec.~\ref{sec:lseg_distill}). 

\subsection{Efficient Depth Map Prediction}
\label{3.1}
The first step of our \ourpaper is to predict the depth map from the given inputs for further initiating the Gaussians. Typical image-to-image depth estimation pipelines like ViT-based encoder-decoder~\cite{dosovitskiy2020image} often lead to noisy edge artifacts in rendering results, caused by directly regressing point maps from image pairs.
Therefore, we instead estimate depth maps for both target and source RGB images through feature matching with cost volumes. This method aggregates feature similarities across views, thereby enhancing the model’s cross-view awareness. These depth maps are then converted to point clouds, upon which Gaussian parameters are regressed. 

\paragraph{Multi-View Feature Extraction.} We employ a CNN to extract down-sampled features from input views. These features are then processed by a Swin-Transformer~\cite{liu2022swin, xu2022gmflow, xu2023unifying},  equipped with cross-attention layers to propagate information across views, enhancing the model’s ability to capture inter-view relationships.  The resulting multi-view-aware features are represented as ${\left\{F_i\in{\mathbb{R}^{\frac{H}{s}\times{\frac{W}{s}}\times{C}}}\right\}_{i=1}^N}$, where $s$ is the down sampling factor and $C$ is the feature dimension. Cross-attention is applied bidirectionally across all views, generalizing the framework to arbitrary numbers of input images. 
\paragraph{Feature Matching and Depth Regression.} Following MVSplat~\cite{mvsplat}, we adopt a plane-sweep stereo~\cite{xu2023unifying, collins1996space} approach for cross-view feature matching. For each view $i$ , we uniformly sample $D$ depth candidates $\left\{d_m\right\}_{m=1}^D$ from the near-to-far depth range. Features from another view $j$  are warped to view $i$ at each depth candidate $d_m$, generating $D$ warped features $\left\{F_{d_m}^{j\to{i}}\right\}_{m=1}^D$  The correlation between these warped features and view $i$ 's original features is computed to construct the cost volume $\left\{C_i \in{\mathbb{R}^{\frac{H}{s}\times{\frac{W}{s}}\times{D}}}\right\}$. Finally, a 2D U-Net with a softmax layer predicts the per-view depth map by processing the concatenated Transformer features and cost volumes. 

\subsection{Multi-cond Semantic Features Aggregation
}\label{3.2}
Recent Works~\cite{chen2024feat2gs} investigate the $\textbf{\textit{geometry awareness}}$  and  $\textbf{\textit{texture awareness}}$ of visual foundation models (VFMs)~\cite{bommasani2021opportunities}, which can enhance scene understanding. Leveraging the capabilities of VFMs, we aggregate pre-trained monocular multi-task semantic features into the cost volume to address challenging scenarios. 
\paragraph{Multi-conditioned Semantic Feature Fusion.} We leverage the pre-trained segmentation backbone from the Segment Anything Model (SAM)~\cite{SAM}and CLIP-LSeg model~\cite{lseg} to get monocular features for each view.  By interpolating to align with the cost volume resolution (\textbf{Sec.\ref{3.1}}), the processed segmentation-semantic features  ${\left\{F{_{i}^{SAM}}\in{\mathbb{R}^{\frac{H}{s}\times{\frac{W}{s}}\times{C_{SAM}}}}\right\}_{i=1}^N}$ from SAM and language-semantic features  ${\left\{F{_{i}^{LSeg}}\in{\mathbb{R}^{\frac{H}{s}\times{\frac{W}{s}}\times{C_{LSeg}}}}\right\}_{i=1}^N}$  from CLIP-LSeg are concatenated with cost volumes$\left\{C_i \in{\mathbb{R}^{\frac{H}{s}\times{\frac{W}{s}}\times{D}}}\right\}_{i=1}^N$ and processed by a lightweight 2D U-Net~\cite{rombach2022high, ronneberger2015u} to regress a unified latent feature map, integrating geometric and semantic cues.

\subsection{Semantic Anisotropic Gaussians Prediction}
\label{3.3}
Despite significant progress in scene understanding and language-guided reconstruction, existing methods~\cite{LSM, semantic-nerf, Langsplat, li2024langsurf} often exhibit limited holistic scene comprehension and rely on single-modal segmentation. To address this, we propose holistic semantic field reconstruction via disentangled segmentation feature distillation and language feature distillation, implemented through anisotropic semantic Gaussians. 

Compared to conventional Gaussians ${\left\{({\mu_j, \alpha_j,\sum_j,c_j})\right\}_{j=1}^n}$~\cite{3DGS}, semantic anisotropic Gaussians ${\left\{({\mu_j, \alpha_j,\sum_j,c_j,f_j})\right\}_{j=1}^n}$ incorporate latent space $\textbf{f}$ into Gaussian attributes to represent 3D semantic fields, enabling joint rendering of novel-view RGB map $\textbf{C}$ and semantic feature map $\textbf{F}$.
\textbf{\begin{equation}
  \begin{cases}
   \textbf{C}=\sum_{i=1}^n\textbf{c}_i\alpha_iG_i(X)\prod_{i=1}^{i-1}(1-\alpha_jG_j(X)),
   \\
   \textbf{F}=\sum_{i=1}^n\textbf{f}_i\alpha_iG_i(X)\prod_{i=1}^{i-1}(1-\alpha_jG_j(X)),
  \end{cases}
\end{equation}
}
Here $G(X)$ stands for the projected 2D Gaussian kernel evaluated at pixel $X$.

Our pipeline proceeds as follows:
\begin{enumerate}
    \item Initialization: Per-view depth maps from cost volumes are unprojected to 3D point clouds (as Gaussian centers $\mu_j$) using camera parameters.
    \item Attribute Prediction: Standard Gaussian parameters (opacity $\alpha_j$, covariance $\Sigma_j$, color $c_j$) are predicted via two convolutional layers processing concatenated inputs (image features, cost volumes, and multi-view images). And the semantic latent attribute $f_j$ is regressed from different heads with the input as the latent feature map, depth, and view-aware features to match multi-modal segmentation.
\end{enumerate}

Pre-trained visual foundation models (VFMs)~\cite{bommasani2021opportunities} often yield feature maps lacking view consistency and spatial awareness. To address this, we introduce a two-stage semantic feature distillation framework, including Segmentation Feature Field Distillation (Sec.~\ref{sec:sam_distill}) and Hierarchical-Context-Aware Language Field Distillation (Sec.~\ref{sec:lseg_distill}) that lifts 2D features to 3D and refines latent feature maps, integrating diverse feature fields for holistic scene understanding.

\subsection{Segmentation Feature Field}
\label{sec:sam_distill}
We leverage the Segment Anything Model (SAM)~\cite{SAM}—an advanced promptable segmentation model supporting inputs like points and bounding boxes—to distill segmentation-semantic embeddings into anisotropic Gaussians. 

\paragraph{Feature Alignment.} From the Gaussian semantic latent attribute $f_j$, an additional segmentation-semantic head is introduced to predict the segmentation-semantic \textbf{$f_j^S$}, as show in Figure~\ref{fig:pipeline}. After rasterization to 2D, we minimize the cosine similarity between the rasterized segmentation feature maps $S={\left\{S_i\in{\mathbb{R}^{h^{'}\times{w^{'}}\times{d^{'}}}}\right\}_{i=1}^N}$, and the SAM encoder outputs $\hat{S}={\left\{\hat{S}_i\in{\mathbb{R}^{H^{'}\times{W^{'}}\times{C^{'}}}}\right\}_{i=1}^N}$. To improve efficiency and reduce memory consumption, we obtain compressed feature maps that are then upsampled to match the SAM features using CNN.

\textbf{\begin{equation}
L_{dist}^{Seg} = 1- \texttt{sim}(f_{expand}(S), \hat{S})=1-\frac{S\cdot{\hat{S}}}{||S||\cdot{||\hat{S}||}}
\end{equation}
}

\paragraph{Prompt-Aware Mask Refinement.} We integrate SAM’s pre-trained mask decoder into our pipeline to generate segmentation masks. A consistency loss enforces alignment between masks derived from our image embeddings $M={\left\{M_i\in{\mathbb{R}^{H\times{W}}}\right\}_{i=1}^M}$ and SAM embeddings $\hat{M}={\left\{\hat{M}_i\in{\mathbb{R}^{H\times{W}}}\right\}_{i=1}^M}$, ensuring promptable segmentation compatibility. We employ a linear combination of Focal Loss~\cite{lin2017focal} and Dice Loss~\cite{milletari2016v} in a 20:1 ratio for optimization.

Focal Loss: \begin{equation}
FL(p_{t} )=-\alpha (1-p_{t})^{\gamma } log(p_{t})
\end{equation}

Dice Loss: \begin{equation}
d=1-\frac{2\left | X\cap Y \right | } {\left | X \right |  + \left | Y \right |  } 
\end{equation}

\textbf{\begin{equation}
L_{mask}^{Seg}=FL(p_{t})+\frac{1}{20} d
\end{equation}
}

\subsection{Hierarchical-Context-Aware Language Field}
\label{sec:lseg_distill}
We leverage CLIP-LSeg—a language-driven segmentation model that aligns textual descriptions with visual content via CLIP embeddings—to distill language-semantic embeddings into anisotropic Gaussians. 

\paragraph{Feature Alignment.} With the segmentation feature branch (Sec.~\ref{sec:sam_distill}) frozen for stability, we use a new language-semantic head to predict the language-semantic features \textbf{$f_j^L$} from the Gaussian semantic latent attribute $f_j$, as show in Figure~\ref{fig:pipeline}, and then rasterize it into 2D language feature maps $L={\left\{L_i\in{\mathbb{R}^{h^{''}\times{w^{''}}\times{d^{''}}}}\right\}_{i=1}^N}$ and expand it to the dimension of CLIP-LSeg~\cite{lseg} feature maps $\hat{L}={\left\{\hat{L_i}\in{\mathbb{R}^{H^{''}\times{W^{''}}\times{C^{''}}}}\right\}_{i=1}^N}$ as $\overline{L}$ for loss computation.
\textbf{\begin{equation}
L_{dist}^{Lang} = 1- \texttt{sim}(\overline{L}, \hat{L})=1-\frac{\overline{L}\cdot{\hat{L}}}{||\overline{L}||\cdot{||\hat{L}||}}
\end{equation}
}
\paragraph{Hierarchical-Context-Aware Pooling.} We employ hierarchical-mask pooling on our expanded language-semantic features $\overline{L}$
to enable fine-grained segmentation (e.g., object parts, materials): here we use SAM to extract three-scale masks ${\left\{\mathbb{M}^h=\left\{m_j^h\right\}_{j=1}^K\right\}_{h=s, m, l}}$ (small/medium/large) to capture hierarchical object contexts. For each scale, L-Seg features within SAM-generated masks are aggregated via average pooling, enhancing intra-mask semantic consistency.
\textbf{\begin{equation}
\overline{L}^h=\frac{\sum\overline{L}\cdot{M^h}}{\sum{M^h}}
\end{equation}
}

\subsection{Training Loss}
During training, our model optimizes 3D anisotropic Gaussians $\{(\mu_j, \alpha_j, \Sigma_j, c_j, f_j)\}_{j=1}^{H \times W \times N}$ through a \textit{two-stage loss formulation} with a combination of photometric loss and feature distillation loss that jointly enforces photometric fidelity and semantic consistency.  

\paragraph{Photometric Loss.}
\begin{equation}
\mathcal{L}_{\text{rgb}} = \sum_{i} \|\mathcal{R_C}(\mu, \alpha, \Sigma, c, f)_i - I^{\text{gt}}_i\|_1 + \lambda_1 \cdot \text{LPIPS}(\mathcal{R}_i, I^{\text{gt}}_i)
\end{equation}
where $\mathcal{R_C}$ is the differentiable renderer of image and $I^{\text{gt}}$ the target image. And the loss weights of LPIPS~\cite{zhang2018lpips} loss weight $\lambda_1$ is set to 0.05.

\paragraph{Semantic Distillation Loss}
(SAM Alignment):  
\begin{equation}
\mathcal{L}_{\text{sam}}=L_{dist}^{Seg}+\lambda_{mask}L_{mask}^{Seg}
\end{equation}
where $\lambda_{mask}$ is set to 0.2.

\paragraph{Hierarchical-Context-Aware Distillation Loss} (CLIP-LSeg Alignment, with segmentation feature branch frozen):  
\begin{equation}
\mathcal{L}_{\text{clip}}=L_{dist}^{Lang}
\end{equation}





\section{Experiments}
\subsection{Settings}

\paragraph{Datasets.} We utilize the ScanNet~\cite{scannet} dataset for training, which provides high-fidelity 3D geometry and high-resolution RGB images, along with estimated camera intrinsic and extrinsic parameters for each frame. A total of 1,462 scenes are used for training, while 50 unseen scenes are reserved for validation. All frames are cropped and resized to a resolution of 256 × 256. 

\paragraph{Metrics.} To evaluate photometric fidelity, we adopt standard image quality metrics: pixel-level PSNR, patch-level SSIM~\cite{wang2004ssim}, and feature-level LPIPS~\cite{zhang2018lpips}. For semantic segmentation, we measure performance using mean Intersection-over-Union (mIoU) and mean pixel accuracy (mAcc). 

\paragraph{Implementation details} Our model is trained using Adam~\cite{kingma2014adam} optimizer with an initial learning rate of $1e-4$ and cosine decay following. Both semantic fea-
ture distillation stages are trained on 4 Nvidia A100 GPU for 5000 iterations.

\begin{figure*}[!t]
\centering
 \begin{minipage}{0.072\linewidth}
    \centerline{Input views}
    \centering
    \begin{minipage}{1\linewidth}
        \centering
        \centerline{\includegraphics[width=\textwidth]{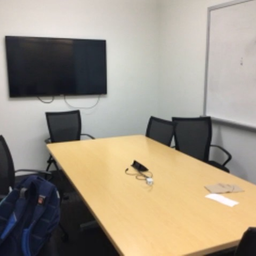}}
     	\centerline{\includegraphics[width=\textwidth]{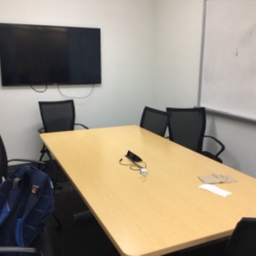}}
        \centerline{\includegraphics[width=\textwidth]{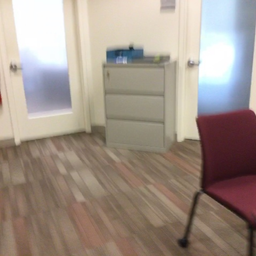}}
     	\centerline{\includegraphics[width=\textwidth]{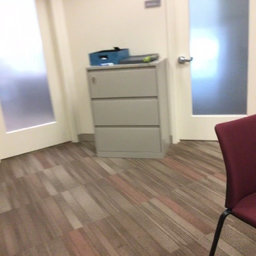}}
        \centerline{\includegraphics[width=\textwidth]{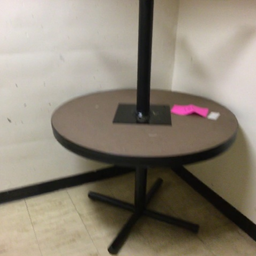}}
     	\centerline{\includegraphics[width=\textwidth]{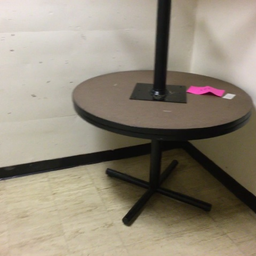}}
        \centerline{\includegraphics[width=\textwidth]{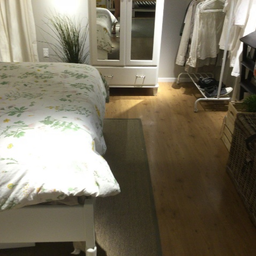}}
     	\centerline{\includegraphics[width=\textwidth]{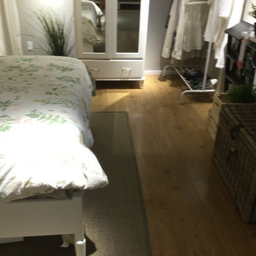}}
    \end{minipage}
 \end{minipage}
 \begin{minipage}{0.145\linewidth}
 \centerline{GT\vphantom{g}}
    \centering
	\centerline{\includegraphics[width=\textwidth]{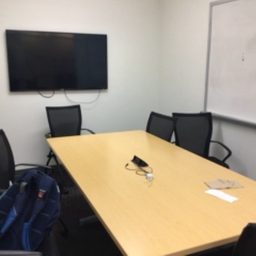}}
	\centerline{\includegraphics[width=\textwidth]{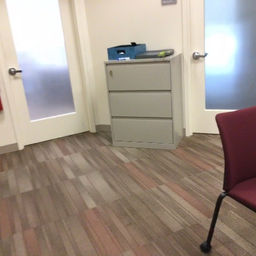}}
    \centerline{\includegraphics[width=\textwidth]{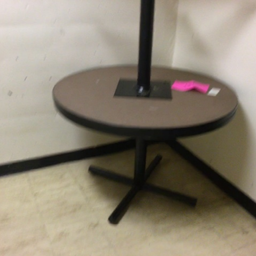}}
	\centerline{\includegraphics[width=\textwidth]{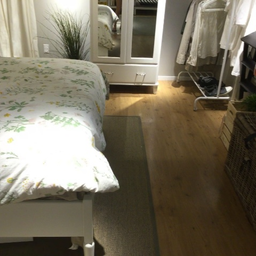}}
\end{minipage}
\begin{minipage}{0.145\linewidth}
\centerline{Ours\vphantom{g}}
    \centering
	\centerline{\includegraphics[width=\textwidth]{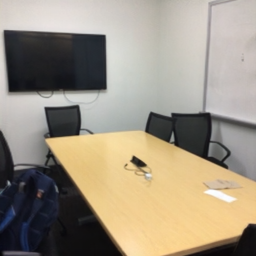}}
	\centerline{\includegraphics[width=\textwidth]{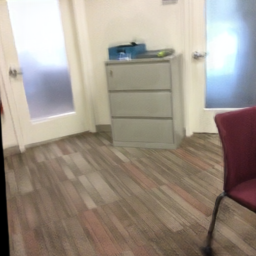}}
    \centerline{\includegraphics[width=\textwidth]{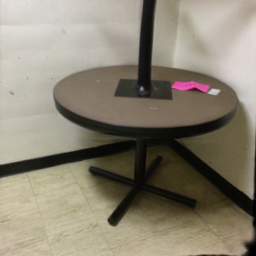}}
	\centerline{\includegraphics[width=\textwidth]{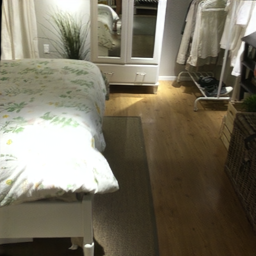}}
\end{minipage}
\begin{minipage}{0.145\linewidth}
\centerline{LSM\vphantom{g}}
    \centering
	\centerline{\includegraphics[width=\textwidth]{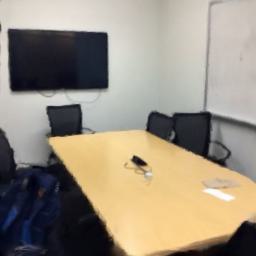}}
	\centerline{\includegraphics[width=\textwidth]{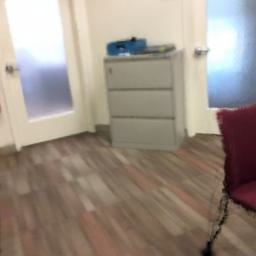}}
    \centerline{\includegraphics[width=\textwidth]{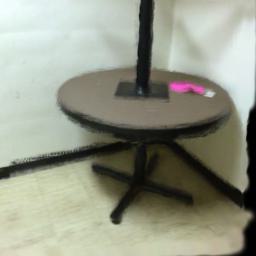}}
	\centerline{\includegraphics[width=\textwidth]{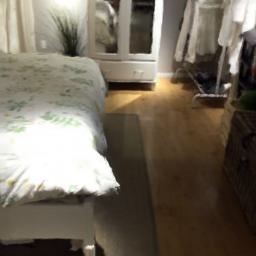}}
\end{minipage}
\begin{minipage}{0.145\linewidth}
\centerline{Feat-3DGS-LSeg}
    \centering
	\centerline{\includegraphics[width=\textwidth]{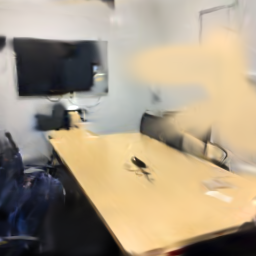}}
	\centerline{\includegraphics[width=\textwidth]{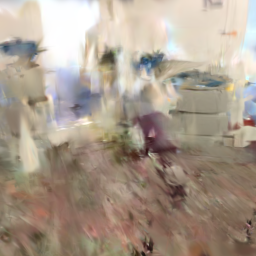}}
    \centerline{\includegraphics[width=\textwidth]{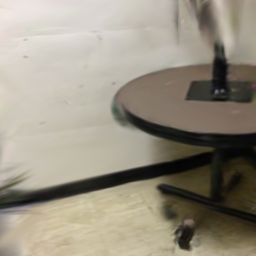}}
	\centerline{\includegraphics[width=\textwidth]{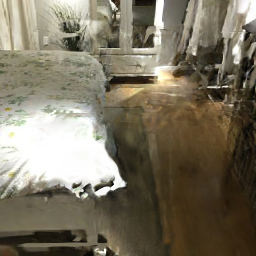}}
\end{minipage}
\begin{minipage}{0.145\linewidth}
\centerline{Feat-3DGS-SAM\vphantom{g}}
    \centering
	\centerline{\includegraphics[width=\textwidth]{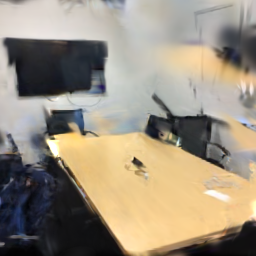}}
	\centerline{\includegraphics[width=\textwidth]{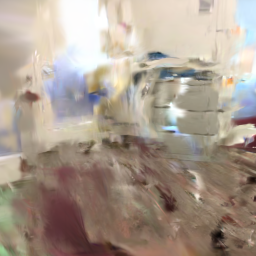}}
    \centerline{\includegraphics[width=\textwidth]{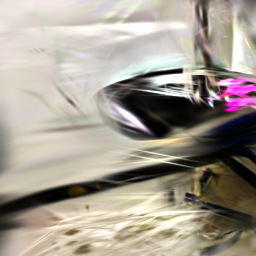}}
	\centerline{\includegraphics[width=\textwidth]{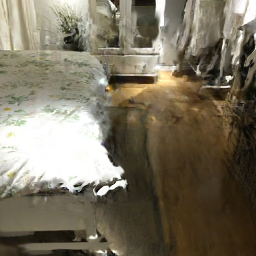}}
\end{minipage}
\begin{minipage}{0.145\linewidth}
\centerline{MVSplat\vphantom{g}}
    \centering
	\centerline{\includegraphics[width=\textwidth]{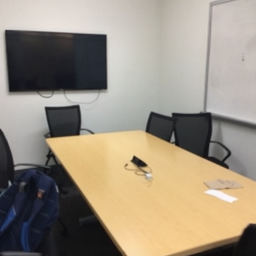}}
	\centerline{\includegraphics[width=\textwidth]{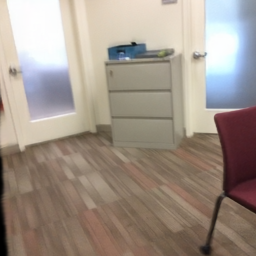}}
    \centerline{\includegraphics[width=\textwidth]{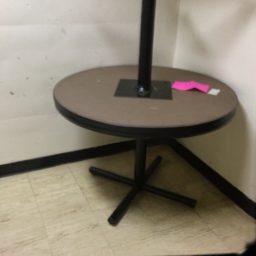}}
	\centerline{\includegraphics[width=\textwidth]{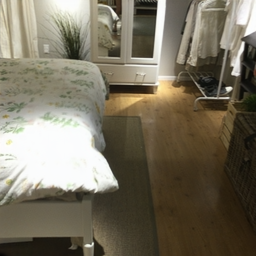}}
\end{minipage}
\caption{\textbf{Novel View Synthesis Comparisons.} Our method outperforms LSM and Feature-3DGS in challenging regions and is compatible with baseline MVSplat, which shows we reconstruct the appearance successfully}
\label{fig:rgb_com}
\end{figure*}

\subsection{Holistic Semantic Field Reconstruction}

\begin{figure*}[!t]
\label{rgb_com}
\centering
 \begin{minipage}{0.15\linewidth}
 \centerline{RGB\vphantom{g}}
    \centering
	\centerline{\includegraphics[width=\textwidth]{figures/images/scene705/000010_gt.png}}
	\centerline{\includegraphics[width=\textwidth]{figures/images/scene691/000250_gt.png}}
    \centerline{\includegraphics[width=\textwidth]{figures/images/scene703/000040_gt.png}}
	\centerline{\includegraphics[width=\textwidth]{figures/images/scene706/000010_gt.png}}
\end{minipage}
 \begin{minipage}{0.15\linewidth}
 \centerline{GT\vphantom{g}}
    \centering
	\centerline{\includegraphics[width=\textwidth]{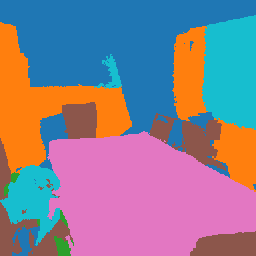}}
	\centerline{\includegraphics[width=\textwidth]{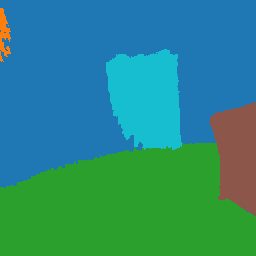}}
    \centerline{\includegraphics[width=\textwidth]{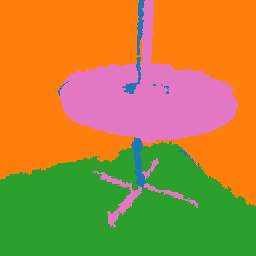}}
	\centerline{\includegraphics[width=\textwidth]{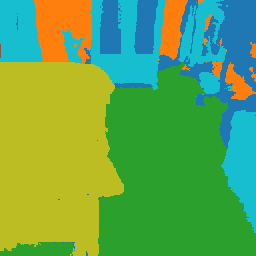}}
\end{minipage}
 \begin{minipage}{0.15\linewidth}
 \centerline{LSeg\vphantom{g}}
    \centering
	\centerline{\includegraphics[width=\textwidth]{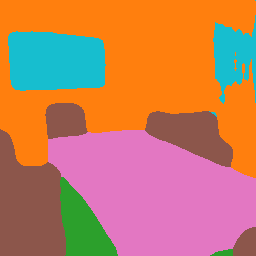}}
	\centerline{\includegraphics[width=\textwidth]{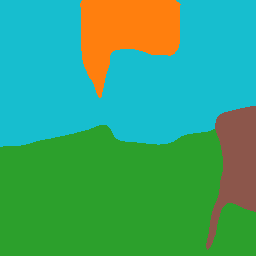}}
    \centerline{\includegraphics[width=\textwidth]{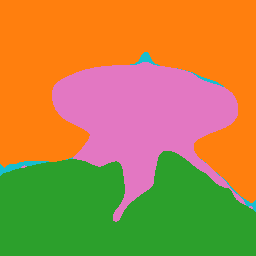}}
	\centerline{\includegraphics[width=\textwidth]{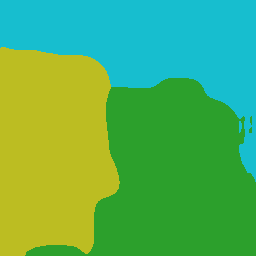}}
\end{minipage}
\begin{minipage}{0.15\linewidth}
\centerline{Ours\vphantom{g}}
    \centering
	\centerline{\includegraphics[width=\textwidth]{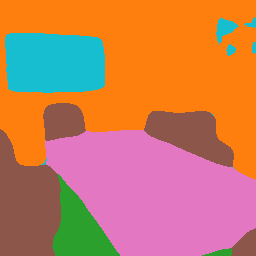}}
	\centerline{\includegraphics[width=\textwidth]{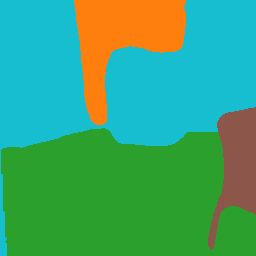}}
    \centerline{\includegraphics[width=\textwidth]{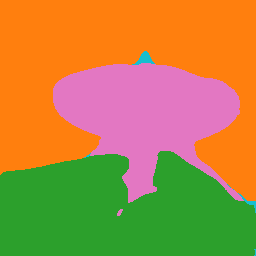}}
	\centerline{\includegraphics[width=\textwidth]{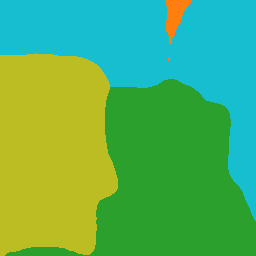}}
\end{minipage}
 \begin{minipage}{0.15\linewidth}
 \centerline{LSM\vphantom{g}}
    \centering
	\centerline{\includegraphics[width=\textwidth]{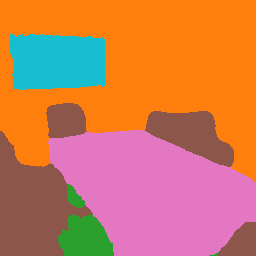}}
	\centerline{\includegraphics[width=\textwidth]{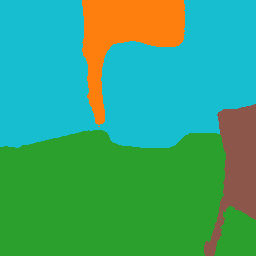}}
    \centerline{\includegraphics[width=\textwidth]{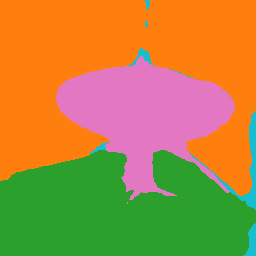}}
	\centerline{\includegraphics[width=\textwidth]{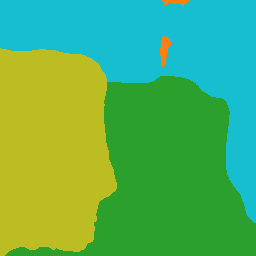}}
\end{minipage}
 \begin{minipage}{0.15\linewidth}
 \centerline{Feature-3DGS\vphantom{g}}
    \centering
	\centerline{\includegraphics[width=\textwidth]{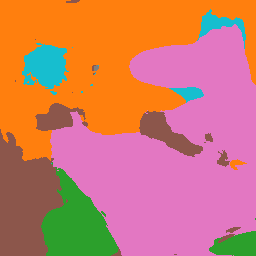}}
	\centerline{\includegraphics[width=\textwidth]{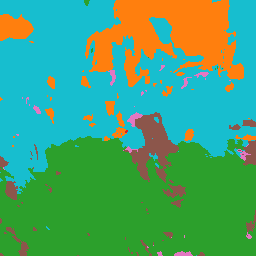}}
    \centerline{\includegraphics[width=\textwidth]{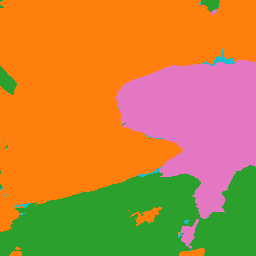}}
	\centerline{\includegraphics[width=\textwidth]{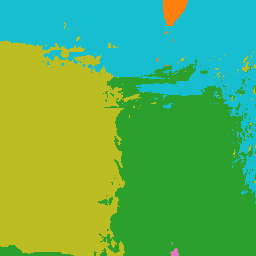}}
\end{minipage}
\qquad
\begin{minipage}{0.92\linewidth}
\centering
\centerline{\includegraphics[width=\textwidth]{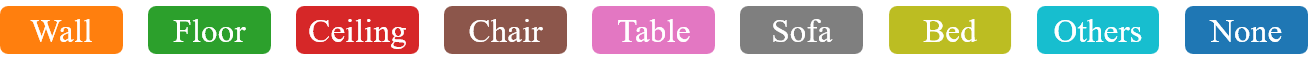}}
\end{minipage}
\caption{\textbf{Language-based Segmentation Comparison.} We visualize the segmentation from a set of categories for unseen view, our method outperforms with the other 3D method and comparably to the 2D VFMs, which indicates we effictively lift 2d foundation language-image model to 3D.}
\label{fig:lang-seg com}
\end{figure*}

\begin{table}[!t]
  \begin{center}
  \footnotesize
  \caption{\textbf{Comparison (Language-Seg)}. Performance metrics for source and target view segmentation across different methods.}
  \setlength{\tabcolsep}{2.2pt}
  \resizebox{\linewidth}{!}{%
    \begin{tabular}{lcc|ccccc}
      \toprule
      Method & \multicolumn{2}{c|}{Source View} & \multicolumn{5}{c}{Target View} \\
      \cmidrule(lr){2-3} \cmidrule(lr){4-8}
      & mIoU$\uparrow$ & Acc.$\uparrow$ & mIoU$\uparrow$ & Acc.$\uparrow$ & PSNR$\uparrow$ & SSIM$\uparrow$ & LPIPS$\downarrow$ \\
      \midrule
      MVSplat           & -      & -      & -      & -      & 23.87 & 0.820 & 0.201 \\
      LSeg              & 0.365  & 0.694  & 0.364  & 0.693  & -     & -     & -     \\
      LSM               & 0.347  & 0.679  & 0.347  & 0.679  & 17.84 & 0.630 & 0.372 \\
      Feature-3DGS      & 0.510  & 0.804  & 0.235  & 0.585  & 13.00 & 0.407 & 0.600 \\
      Ours              & 0.376  & 0.707  & 0.371  & 0.702  & 21.88 & 0.879 & 0.191 \\
      Ours w/HCAM       & 0.376  & 0.707  & 0.386  & 0.710  & 21.88 & 0.879 & 0.191 \\
      \bottomrule
    \end{tabular}%
  }
  \label{tab:Language}
  \end{center}
\end{table}


\begin{table}[!t]
  \begin{center}
  \footnotesize
  \caption{\textbf{Comparison (Promtable-Seg)}. Performance metrics for source and target view segmentation across different methods.}
  \setlength{\tabcolsep}{2.2pt}
  \resizebox{\linewidth}{!}{%
    \begin{tabular}{lcc|ccccc}
      \toprule
      Method & \multicolumn{2}{c|}{Source View} & \multicolumn{5}{c}{Target View} \\
      \cmidrule(lr){2-3} \cmidrule(lr){4-8}
      & mIoU$\uparrow$ & Acc.$\uparrow$ & mIoU$\uparrow$ & Acc.$\uparrow$ & PSNR$\uparrow$ & SSIM$\uparrow$ & LPIPS$\downarrow$ \\
      \midrule
      SAM              & 0.684 & 0.427 & 0.426 & 0.684 & -     & -     & -     \\
      Feature-3DGS     & 0.678 & 0.409 & 0.395 & 0.681 & 13.14 & 0.405 & 0.601 \\
      Ours             & 0.691 & 0.438 & 0.433 & 0.690 & 21.88 & 0.879 & 0.191 \\
      \bottomrule
    \end{tabular}%
  }
  \label{tab:Promtable}
  \end{center}
\end{table}

We compare our approach with two state-of-the-art methods: LSM~\cite{LSM} (a generalizable framework) and Feature-3DGS~\cite{Feature_3dgs} (a per-scene optimization-based method). Both methods predict RGB values and leverage feature-based 3D Gaussian Splatting (3D-GS)~\cite{3DGS}. Unlike our approach, Feature-3DGS supports promptable segmentation and open-vocabulary segmentation by separately optimizing SAM and LSeg features, while LSM is limited to open-vocabulary segmentation.  

\paragraph{Evaluation of Novel View Synthesis}
We further compare our method with the baseline MVSplat~\cite{mvsplat}, a feed-forward Gaussian reconstruction model trained on the RealEstate10K~\cite{real10k} dataset. As shown in Table \ref{tab:Language} and Figure \ref{fig:rgb_com}, Feature-3DGS~\cite{Feature_3dgs} struggles to synthesize high-quality images from sparse input views, while LSM~\cite{LSM} introduces noisy artifacts, especially near object boundaries. Our method achieves comparable pixel-level quality to MVSplat despite training on lower-quality data and outperforms it at the patch and feature levels, owing to the semantic priors integrated into our pipeline.

\begin{figure}[!t]
    \centering
    \label{language feature field rendered at novel views}
    \rightline{language feature field rendered at novel views}
    \begin{minipage}{0.18\linewidth}
        \includegraphics[width=\linewidth]{figures/images/scene691/000240.png}
        \end{minipage}\hfill
    \begin{minipage}{0.18\linewidth}
        \includegraphics[width=\linewidth]{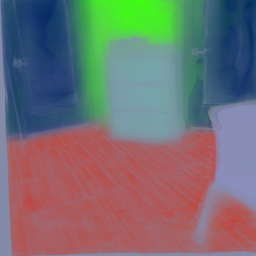}
    \end{minipage}\hfill
    \begin{minipage}{0.18\linewidth}
        \includegraphics[width=\linewidth]{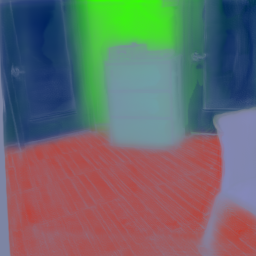}
    \end{minipage}\hfill
    \begin{minipage}{0.18\linewidth}
        \includegraphics[width=\linewidth]{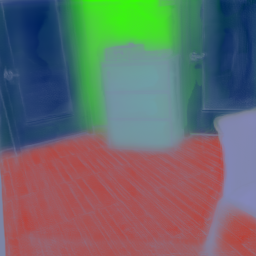}
    \end{minipage}\hfill
    \begin{minipage}{0.18\linewidth}
        \includegraphics[width=\linewidth]{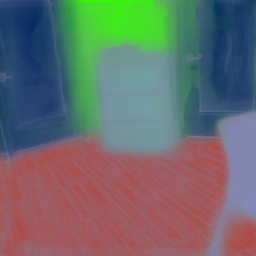}
    \end{minipage}
    
    \vspace{0.1cm} 
    \hspace*{\fill}
    \begin{tikzpicture}
         \draw[thick] (0,0) -- (0.77\linewidth,0); 
    \end{tikzpicture}
    \vspace{0.1cm} 

    \centering
    \begin{minipage}{0.18\linewidth}    
        \includegraphics[width=\linewidth]{figures/images/scene691/000260.png}
        \end{minipage}\hfill
    \begin{minipage}{0.18\linewidth}
        \includegraphics[width=\linewidth]{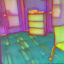}
    \end{minipage}\hfill
    \begin{minipage}{0.18\linewidth}
        \includegraphics[width=\linewidth]{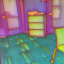}
    \end{minipage}\hfill
    \begin{minipage}{0.18\linewidth}
        \includegraphics[width=\linewidth]{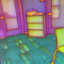}
    \end{minipage}\hfill
    \begin{minipage}{0.18\linewidth}
        \includegraphics[width=\linewidth]{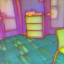}
    \end{minipage}
   \rightline{segmentation feature field rendered at novel views}\hfill
   \caption{\textbf{Visualization of the Semantic Feature Field.} We visualize the language features and segmentation characteristics of the novel views, demonstrating how we elevate the 2D features into 3D while maintaining consistency across views. The visualizations are generated using PCA~\cite{pedregosa2011scikit}.}
   \label{fig:pca}
   
\end{figure}

\paragraph{Evaluation of Open-vocabulary Semantic 3D Segmentation}
Following Feature-3DGS~\cite{Feature_3dgs}, we map thousands of category labels from diverse datasets into a unified set of common categories: \{Wall, Floor, Ceiling, Chair, Table, Bed, Sofa, Others\}. For comparison, we also include the 2D open-vocabulary segmentation method LSeg~\cite{lseg}. As shown in Table \ref{tab:Language} and visualised in Figure \ref{fig:lang-seg com}, our method achieves competitive performance against baseline 3D methods and matches the accuracy of 2D methods when evaluated on the ScanNet dataset with transferred labels. Notably, while LSeg suffers from cross-view inconsistency, our approach maintains high consistency across views. To illustrate this, we visualize the language feature fields of both methods using PCA (projecting high-dimensional features into three channels)~\cite{pca} in Figure \ref{fig:pca}.

\begin{table*}[!t]
  \centering
  \footnotesize
  \caption{\textbf{Feature‑Condition Ablation (Stages 1 \& 2)}.  
    Stage 1: feature branch under LSeg vs. GT masks;  
    Stage 2: feature branch under SAM vs. GT masks.}
  \setlength{\tabcolsep}{2.2pt}
  \resizebox{\linewidth}{!}{%
    \begin{tabular}{lcc|cc|cc|cc}
      \toprule
      Condition
        & \multicolumn{2}{c|}{Compared with LSeg Masks}
        & \multicolumn{2}{c|}{Compared with GT Masks}
        & \multicolumn{2}{c|}{Compared with GT SAM}
        & \multicolumn{2}{c}{Compared with GT Masks} \\
      \cmidrule(lr){2-3} \cmidrule(lr){4-5} \cmidrule(lr){6-7} \cmidrule(lr){8-9}
        & mIoU$\uparrow$ & Acc.$\uparrow$
        & mIoU$\uparrow$ & Acc.$\uparrow$
        & mIoU$\uparrow$ & Acc.$\uparrow$
        & mIoU$\uparrow$ & Acc.$\uparrow$ \\
      \midrule
      full        & 0.630 & 0.894 & 0.368 & 0.701 & 0.668 & 0.847 & 0.433 & 0.690 \\
      SAM         & 0.458 & 0.746 & 0.328 & 0.650 & 0.663 & 0.844 & 0.433 & 0.690 \\
      LSeg        & 0.628 & 0.893 & 0.369 & 0.699 & 0.591 & 0.790 & 0.424 & 0.681 \\
      w/o cond.   & 0.263 & 0.575 & 0.191 & 0.491 & 0.546 & 0.753 & 0.414 & 0.676 \\
      \bottomrule
    \end{tabular}%
  }
  \label{tab:FCA_combined}
\end{table*}

\begin{figure}[!t]
\centering
 \begin{minipage}{0.32\linewidth}
 \centerline{RGB}
    \centering
	\centerline{\includegraphics[width=\textwidth]{figures/images/scene705/000010_gt.png}}
	\centerline{\includegraphics[width=\textwidth]{figures/images/scene691/000250_gt.png}}
    \centerline{\includegraphics[width=\textwidth]{figures/images/scene703/000040_gt.png}}
	\centerline{\includegraphics[width=\textwidth]{figures/images/scene706/000010_gt.png}}
\end{minipage}
 \begin{minipage}{0.32\linewidth}
 \centerline{Ours}
    \centering
	\centerline{\includegraphics[width=\textwidth]{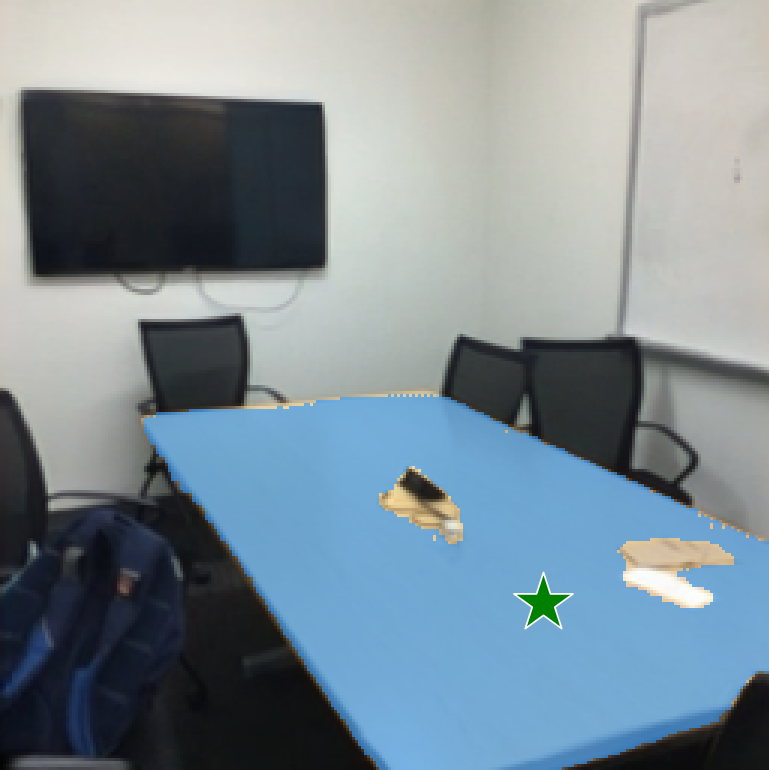}}
	\centerline{\includegraphics[width=\textwidth]{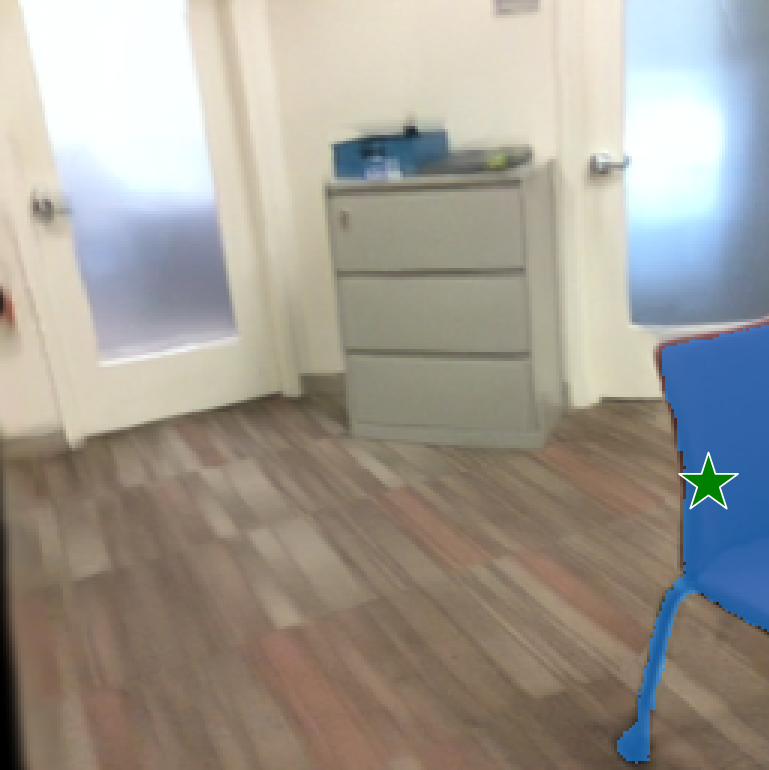}}
    \centerline{\includegraphics[width=\textwidth]{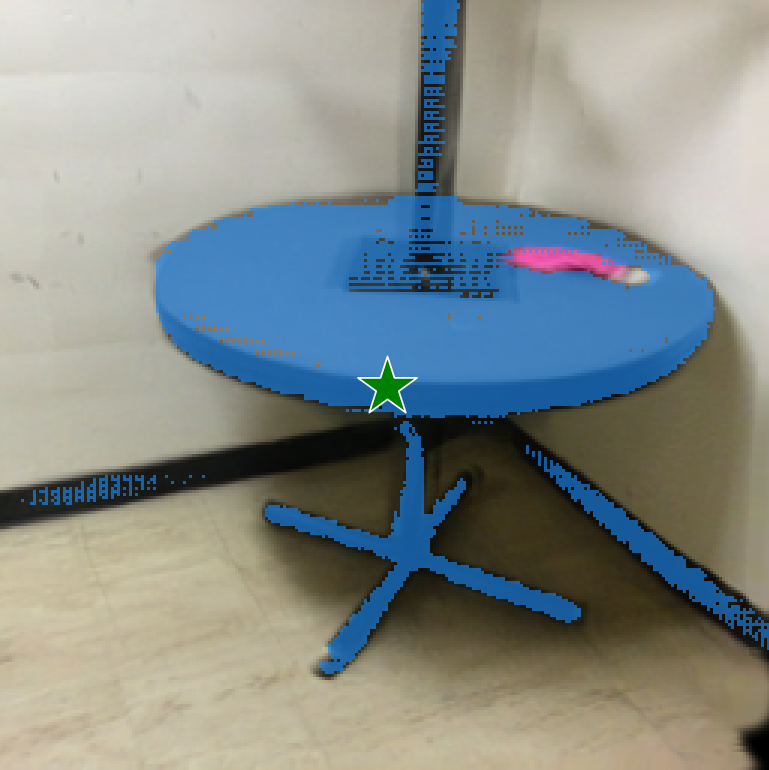}}
	\centerline{\includegraphics[width=\textwidth]{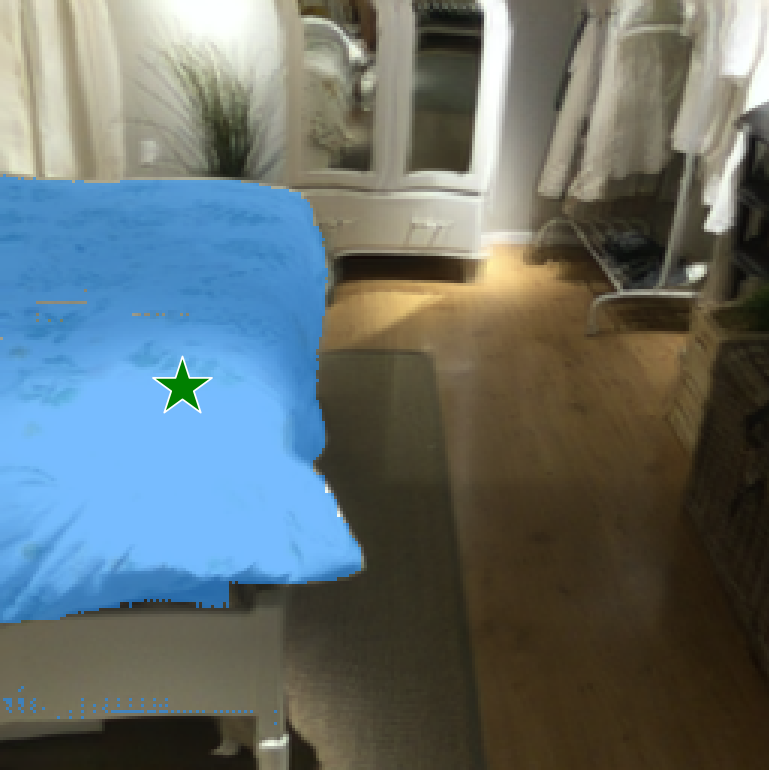}}
\end{minipage}
 \begin{minipage}{0.32\linewidth}
 \centerline{Feature-3DGS}
    \centering
	\centerline{\includegraphics[width=\textwidth]{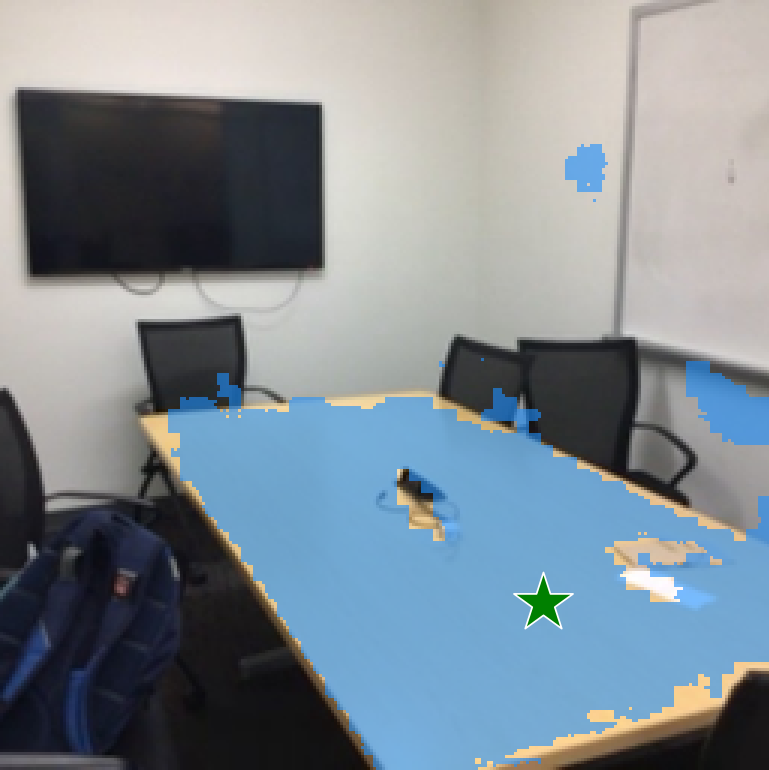}}
	\centerline{\includegraphics[width=\textwidth]{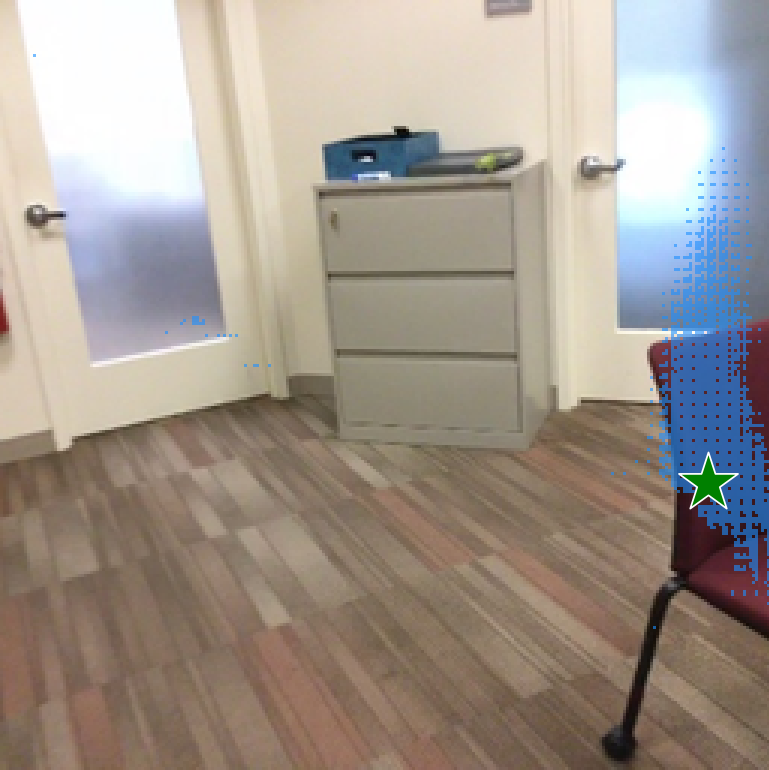}}
    \centerline{\includegraphics[width=\textwidth]{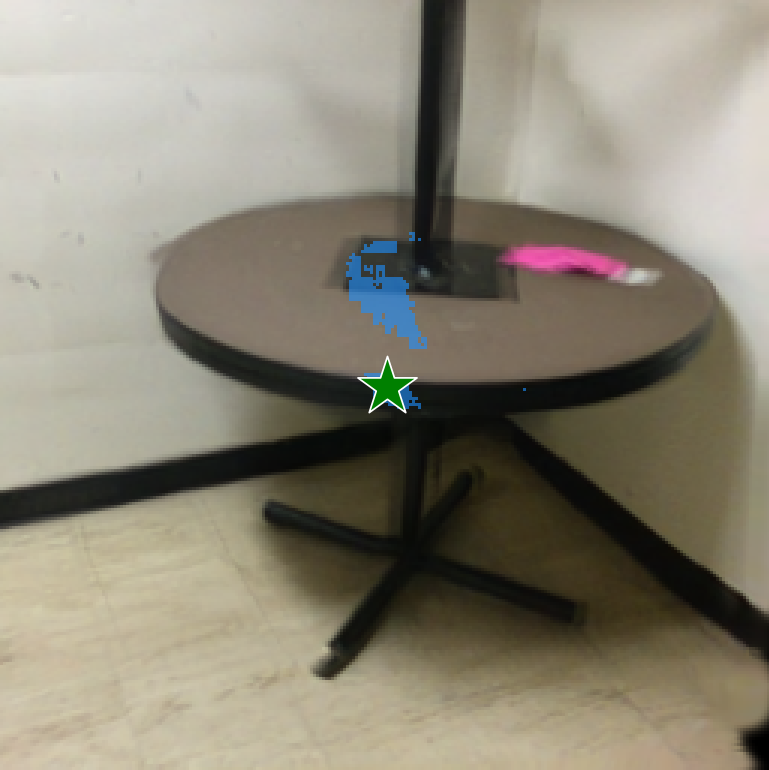}}
    \centerline{\includegraphics[width=\textwidth]{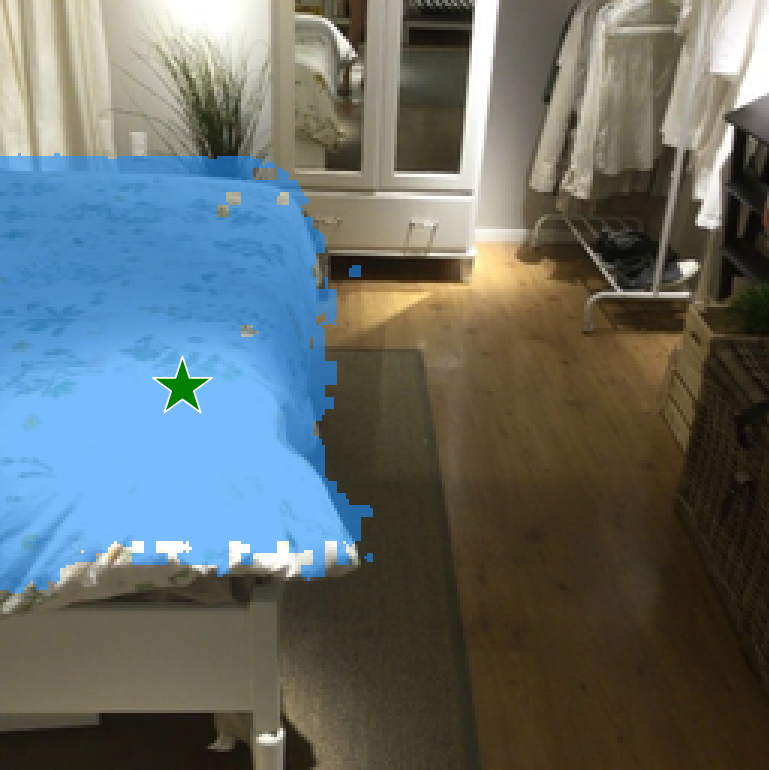}}
\end{minipage}
\caption{\textbf{Prompt-based Segmentation Comparison.} We visualize the segmentation generated from a point prompt for unseen views. Our method outperforms other 3D approaches, indicating that we effectively extend a 2D foundation segmentation model to 3D.}
\label{fig:seg_com}
\end{figure}

\paragraph{Evaluation of Promptable Semantic 3D Segmentation}
Building on the SAM mask decoder, our method predicts three hierarchical masks from point queries and the predicted segmentation feature map. We uniformly sample a grid of points on the images (32 points along the width and 32 along the height, totaling 1,024 points). For each point, we generate hierarchical masks and evaluate their alignment with ground truth masks from ScanNet labels by reporting the highest Intersection-over-Union (IoU) and accuracy (Acc) scores in Table \ref{tab:Promtable} and visualize the promptable segmentation in Figure \ref{fig:seg_com}. In addition to Feature-3DGS, we include the 2D promptable segmenter SAM in our comparisons. To visualize the segmentation feature field and SAM features, we project them into three channels using PCA~\cite{pca} in Figure \ref{fig:pca}.
\subsection{Ablation Studies
}

\begin{figure}[ht]
\label{ablation_language}
\begin{subfigure}{0.49\textwidth}
\centering
 \begin{minipage}{0.15\linewidth}
 \centerline{RGB\vphantom{g}}
    \centering
	\centerline{\includegraphics[width=\textwidth]{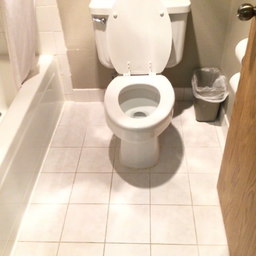}}
	\centerline{\includegraphics[width=\textwidth]{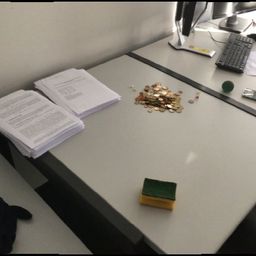}}
\end{minipage}
 \begin{minipage}{0.15\linewidth}
 \centerline{w/o cond.\vphantom{g}}
    \centering
	\centerline{\includegraphics[width=\textwidth]{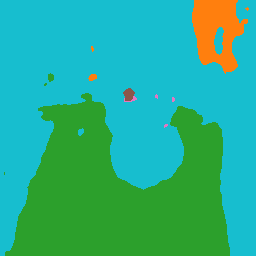}}
	\centerline{\includegraphics[width=\textwidth]{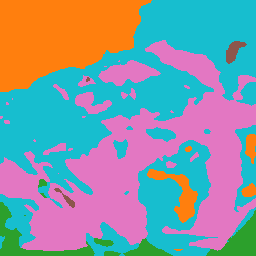}}
\end{minipage}
 \begin{minipage}{0.15\linewidth}
 \centerline{SAM\vphantom{g}}
    \centering
	\centerline{\includegraphics[width=\textwidth]{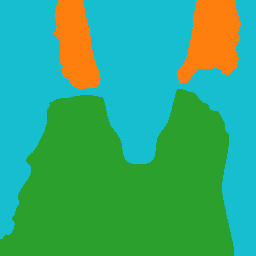}}
	\centerline{\includegraphics[width=\textwidth]{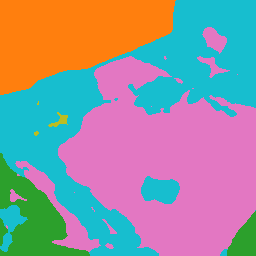}}
\end{minipage}
\begin{minipage}{0.15\linewidth}
\centerline{LSeg}
    \centering
	\centerline{\includegraphics[width=\textwidth]{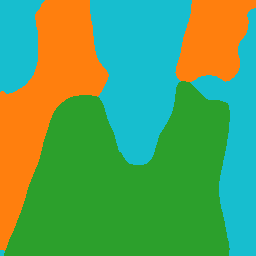}}
	\centerline{\includegraphics[width=\textwidth]{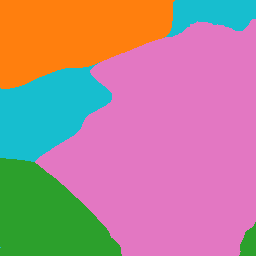}}
\end{minipage}
\begin{minipage}{0.15\linewidth}
\centerline{full\vphantom{g}}
    \centering
	\centerline{\includegraphics[width=\textwidth]{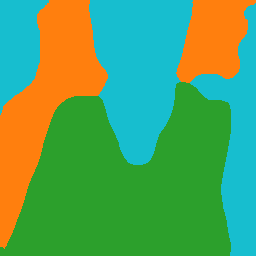}}
	\centerline{\includegraphics[width=\textwidth]{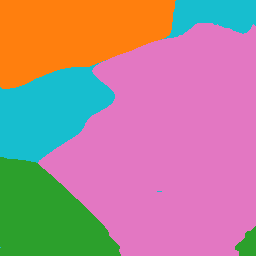}}
\end{minipage}
\begin{minipage}{0.15\linewidth}
\centerline{w/ HCAM\vphantom{g}}
    \centering
	\centerline{\includegraphics[width=\textwidth]{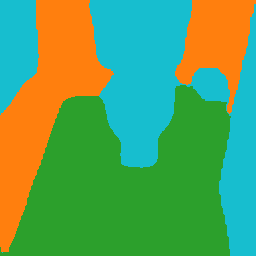}}
	\centerline{\includegraphics[width=\textwidth]{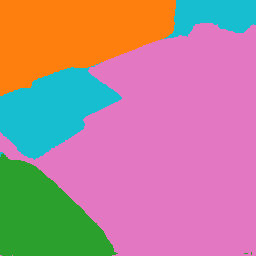}}
\end{minipage}
\caption{Comparisons of language segmentation.}
\label{ablation_lang}
\end{subfigure}
\begin{subfigure}{0.49\textwidth}
\centering
 \begin{minipage}{0.15\linewidth}
 \centerline{RGB\vphantom{g}}
    \centering
	\centerline{\includegraphics[width=\textwidth]{figures/ablation/lseg/color1.png}}
	\centerline{\includegraphics[width=\textwidth]{figures/ablation/lseg/color2.png}}
\end{minipage}
 \begin{minipage}{0.15\linewidth}
 \centerline{w/o cond.\vphantom{g}}
    \centering
	\centerline{\includegraphics[width=\textwidth]{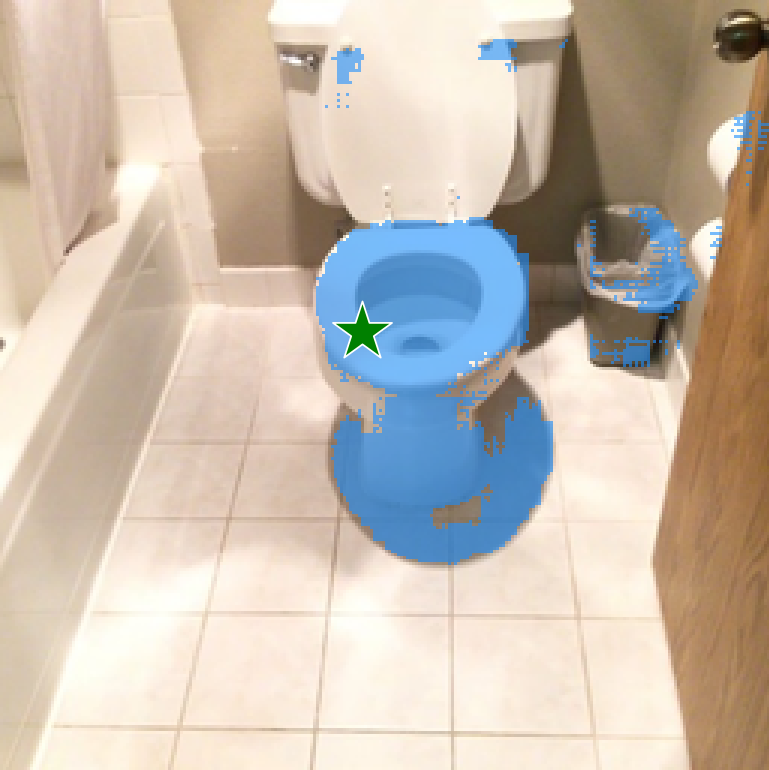}}
	\centerline{\includegraphics[width=\textwidth]{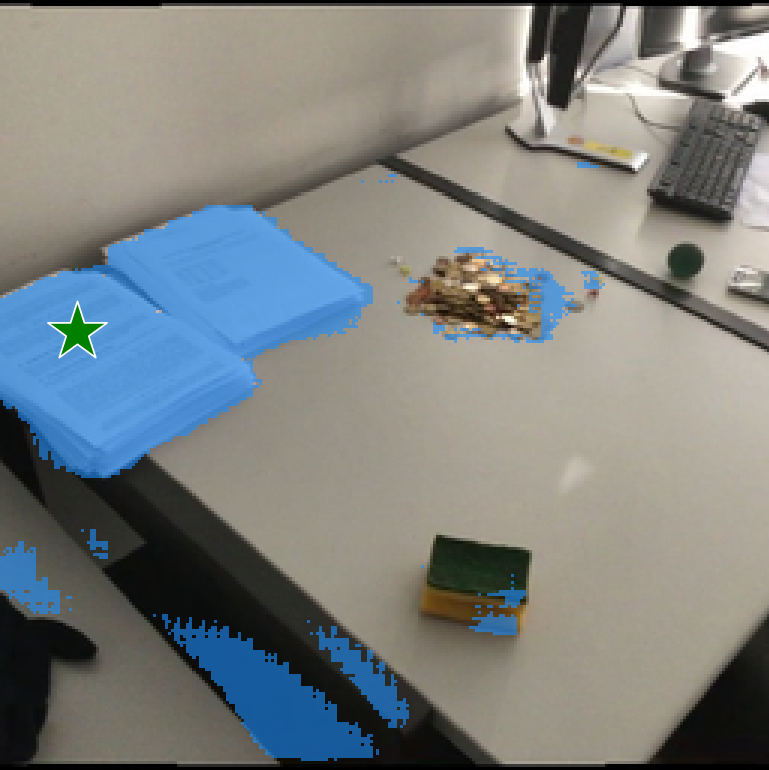}}
\end{minipage}
 \begin{minipage}{0.15\linewidth}
 \centerline{SAM\vphantom{g}}
    \centering
	\centerline{\includegraphics[width=\textwidth]{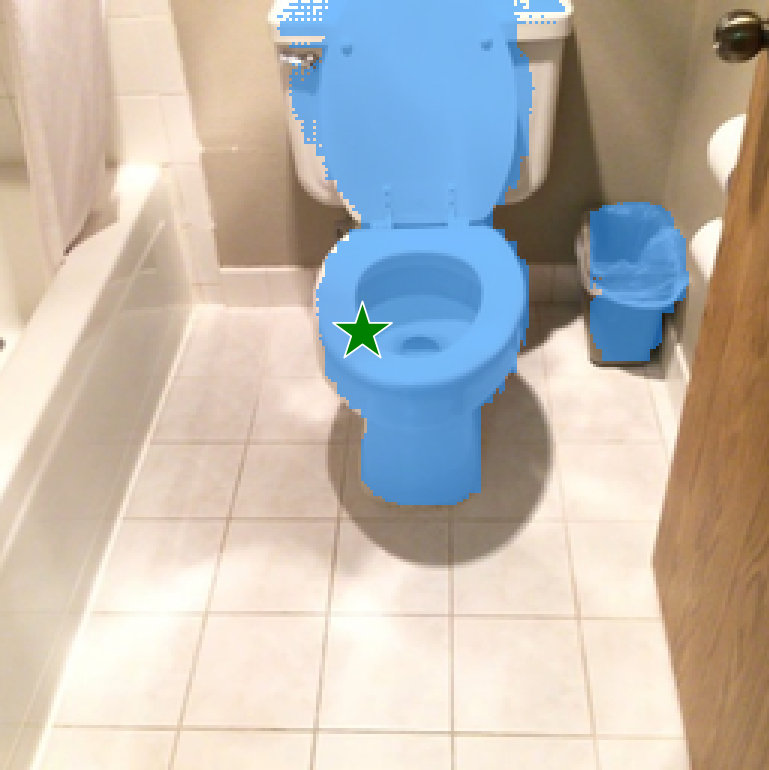}}
	\centerline{\includegraphics[width=\textwidth]{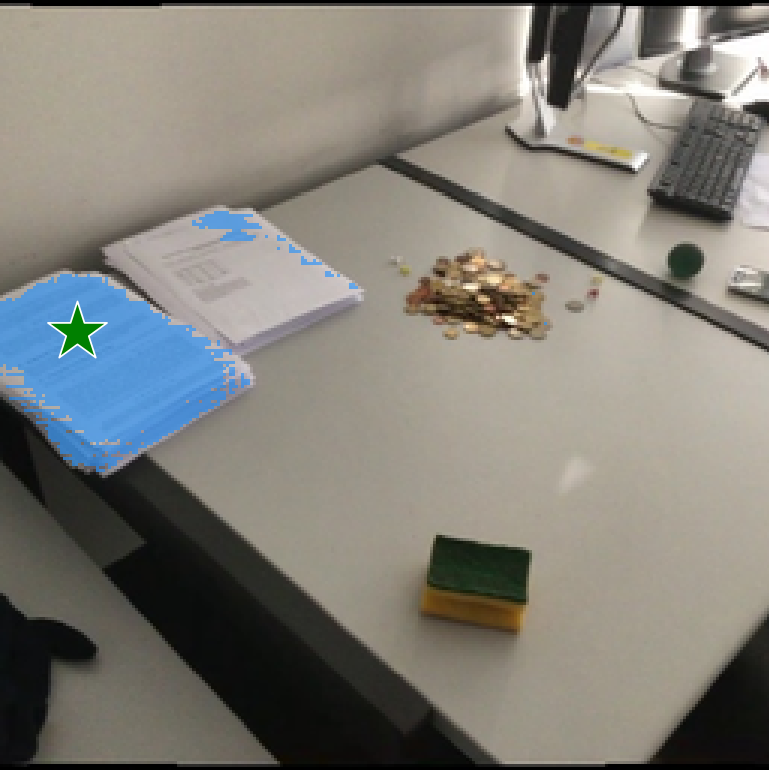}}
\end{minipage}
\begin{minipage}{0.15\linewidth}
\centerline{LSeg\vphantom{g}}
    \centering
	\centerline{\includegraphics[width=\textwidth]{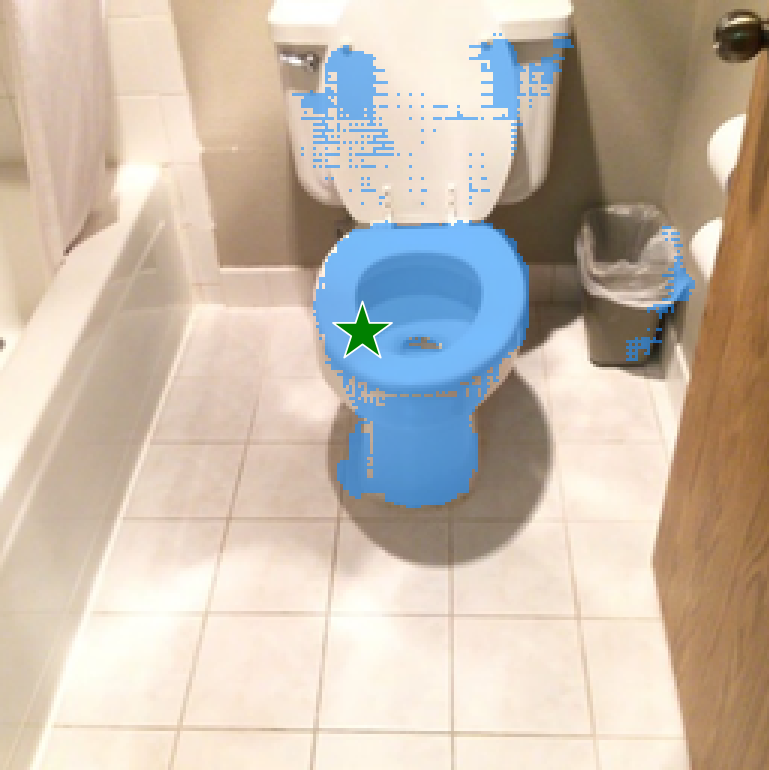}}
	\centerline{\includegraphics[width=\textwidth]{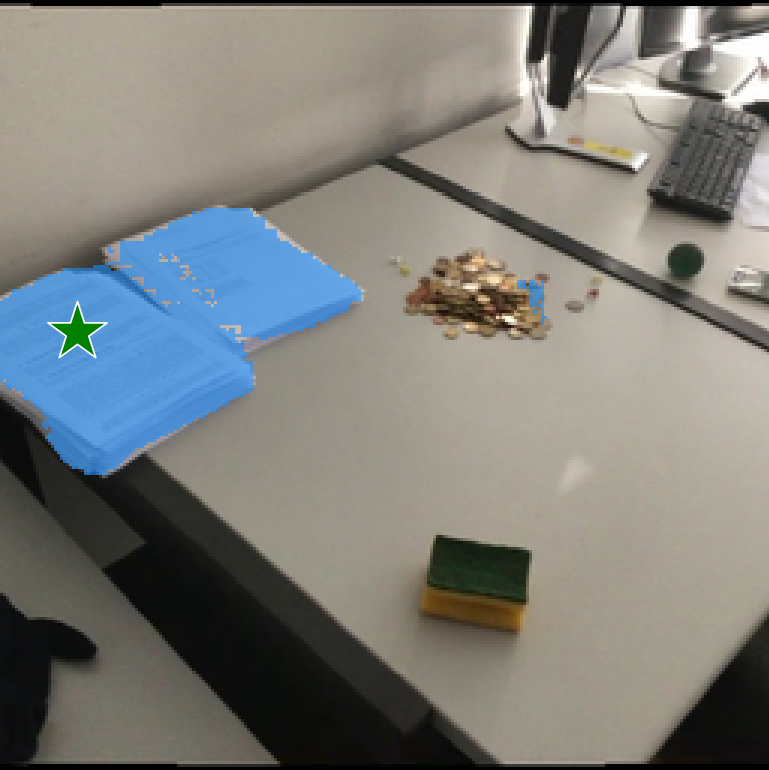}}
\end{minipage}
\begin{minipage}{0.15\linewidth}
\centerline{full\vphantom{g}}
    \centering
	\centerline{\includegraphics[width=\textwidth]{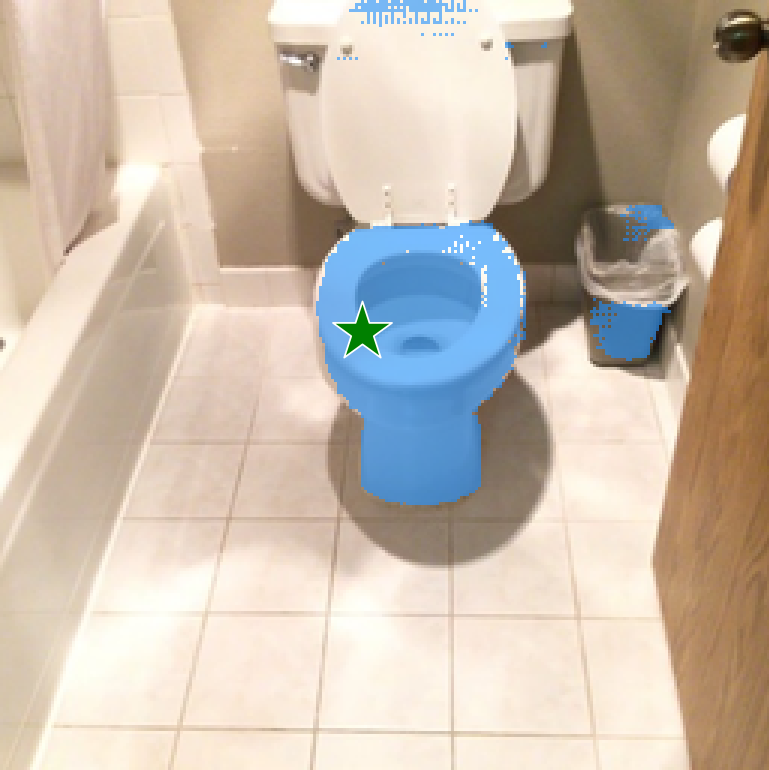}}
	\centerline{\includegraphics[width=\textwidth]{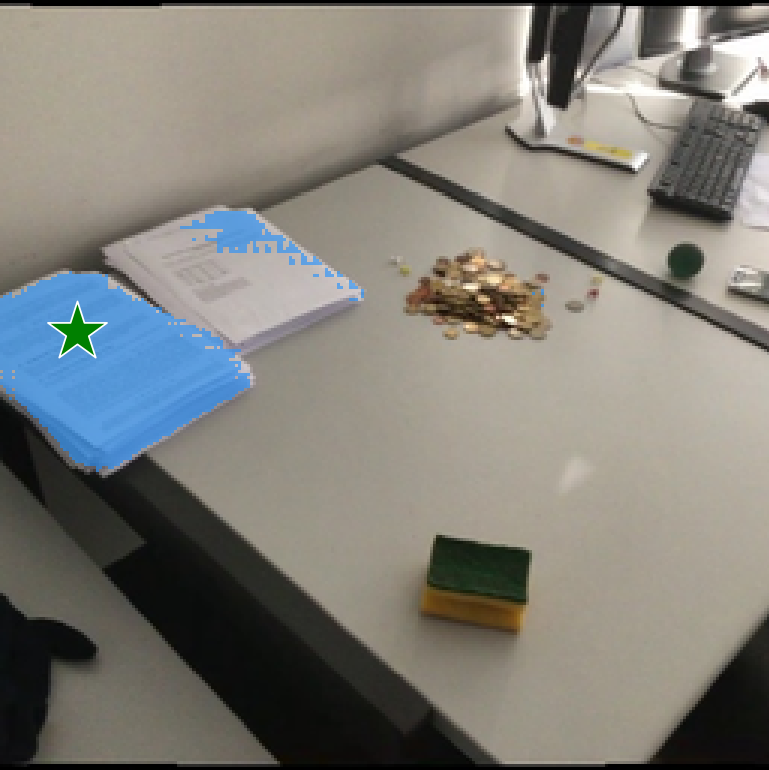}}
\end{minipage}
\begin{minipage}{0.15\linewidth}
\centerline{w/ mask loss\vphantom{g}}
    \centering
	\centerline{\includegraphics[width=\textwidth]{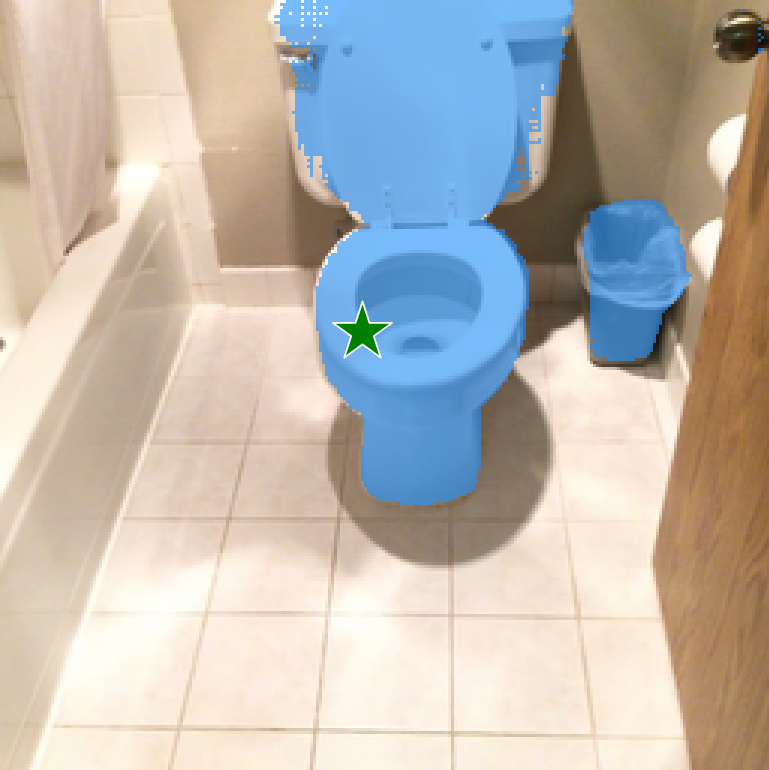}}
	\centerline{\includegraphics[width=\textwidth]{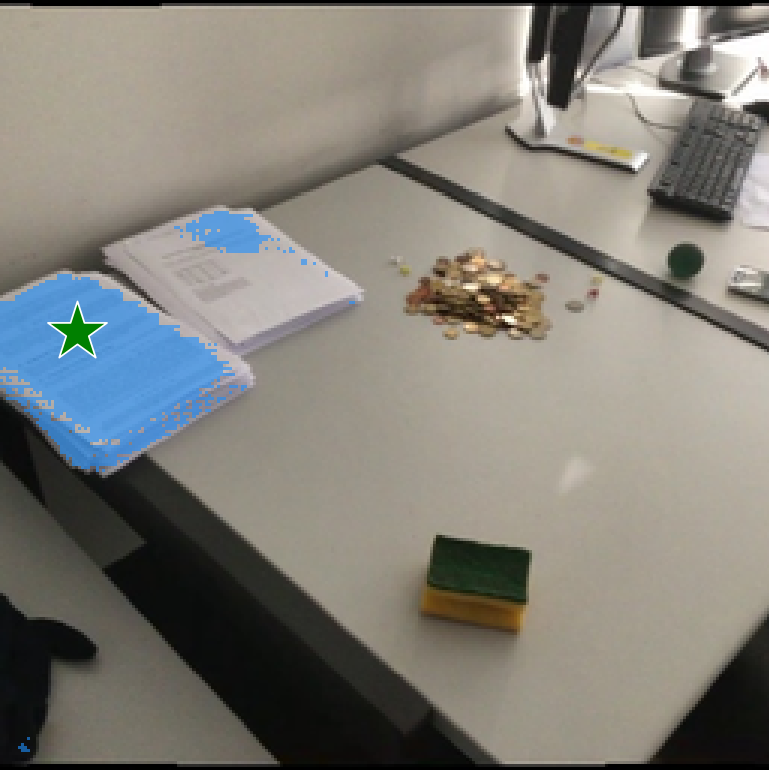}}
\end{minipage}
\caption{Comparisons of promptable segmentation.}
\label{ablation_prompt}
\end{subfigure}
\caption{\textbf{Ablation study on different Conditions, HCAM and Mask loss.} We visualize the segmentation results under different conditions, illustrating that all these are complementary.}
\label{fig:ablation}
\end{figure}




\paragraph{Multi-Conditioned Semantic Features}
In Table \ref{tab:FCA_combined} and Figure \ref{fig:ablation}, we compare our full model with a variant that excludes multi-semantic feature conditioning (Sec. \ref{3.2}), denoted as w/o condition, retaining only the multi-view branches. We further evaluate distinct monocular semantic features: the segmentation-semantic SAM feature and the language-semantic LSeg feature. To assess distillation and segmentation performance, we compare results separately using the 2D feature-derived masks and the ground truth (GT) masks. Our results demonstrate that feature concatenation achieves an optimal balance between promptable and open-vocabulary segmentation performance.

For the ablation study on Mask Loss in Semantic Distillation (Sec.~\ref{sec:sam_distill}), please refer to the supplementary material for more details.

\section{Discussion}

\paragraph{Limitation.} While our method significantly reconstructs the holistic Gaussian feature field, it relies on a pre-trained model for feature lifting, which increases computational and GPU memory requirements. Additionally, our current model requires camera poses as input along with the multi-view images, which could limit scalability for various applications. Future work could explore pose-free models to move the requirement, further bridging the gap between modular design and real-world applicability. 
\paragraph{Conclusion}. In this paper, we propose a novel framework for feature distillation and multi-modal segmentation, leveraging multi-semantic conditioning with Segment Anything Model (SAM) and Language-Semantic (LSeg) features. Our experiments demonstrate that the complete model, incorporating both segmentation-semantic (SAM) and language-semantic (LSeg) features, achieves superior performance in balancing promptable and open-vocabulary segmentation tasks. In conclusion, this work advances the integration of vision-language models into 3D segmentation pipelines, offering a scalable solution for diverse semantic understanding tasks.

{\small
\bibliographystyle{ieeenat_fullname}
\bibliography{11_references}
}

\ifarxiv \clearpage \appendix 
%

\section{Ablation on Mask Loss}
We compare our first-stage feature distillation (full) with a variant that excludes the mask loss (w/o mask loss). We evaluate the segmentation metrics using SAM masks to assess the effectiveness of the distillation process in Table \ref{tab:mask_ablation}.

\begin{table}[!t]
  \centering
  \caption{\textbf{Ablation Study}. Impact of mask loss on segmentation.}
  \setlength{\tabcolsep}{3pt}
  \begin{tabular}{lcc}
    \toprule
    Variant           & mIoU$\uparrow$ & Acc.$\uparrow$ \\
    \midrule
    w/ mask loss      & 0.668          & 0.847 \\
    w/o mask loss     & 0.659          & 0.842 \\
    \bottomrule
  \end{tabular}
  \label{tab:mask_ablation}
\end{table}

\section{Module Timing}
We evaluate the computational cost of each module by running inference on the ScanNet dataset and calculating the runtime for each component of our method, as detailed in Table \ref{tab:runtime_modules}. 

\begin{table}[!t]
  \centering
  \caption{\textbf{Inference Time per Module}.}
  \setlength{\tabcolsep}{3pt}
  \begin{tabular}{lc}
    \toprule
    Module                                                     & Time (s) \\
    \midrule
    Depth map predict w/ condition (3.1+3.2)                   & 0.092 \\
    Segmentation field distillation (stage 1)                  & 0.032 \\
    Language field distillation (stage 2)                      & 0.029 \\
    \midrule
    \textbf{Total}                                             & \textbf{0.153} \\
    \bottomrule
  \end{tabular}
  \label{tab:runtime_modules}
\end{table}


 \fi

\end{document}